\documentclass[journal]{IEEEtran}
\usepackage[numbers]{natbib}
\usepackage{amsmath,amssymb}
\usepackage{graphicx}
\usepackage{float}
\usepackage{tabularx,multirow}
\usepackage{booktabs}
\usepackage{hyperref}
\usepackage{makecell}
\usepackage{arydshln}
\usepackage{pifont}
\usepackage{orcidlink}

\usepackage{xcolor}
\definecolor{darkgreen}{HTML}{2db300}
\definecolor{darkred}{HTML}{AA1803}
\definecolor{codegreen}{HTML}{527A6A}
\definecolor{codepink}{HTML}{DB5CA3}
\definecolor{textgreen}{HTML}{009900}
\definecolor{textorange}{HTML}{ff8000}

\hypersetup{
colorlinks=true,
linkcolor=black,
citecolor=black
}

\begin{document}
\title{SSLChange: A Self-supervised Change Detection Framework Based on Domain Adaptation}

\author{
	Yitao Zhao \hspace{-2mm}$^{~\orcidlink{0000-0001-6605-1757}}$, 
        Turgay Celik \hspace{-2mm}$^{~\orcidlink{0000-0001-6925-6010}}$, 
        \IEEEmembership{Senior Member, IEEE},
        Nanqing Liu \hspace{-2mm}$^{~\orcidlink{0000-0001-7564-4896}}$,
        \IEEEmembership{Graduate Student Member, IEEE}, \\
        Feng Gao \hspace{-2mm}$^{~\orcidlink{0000-0002-1825-328X}}$,
        \IEEEmembership{Member, IEEE}, 
        and Heng-Chao Li$^{*}$ \hspace{-2mm}$^{~\orcidlink{0000-0002-9735-570X}}$,
        \IEEEmembership{Senior Member, IEEE}

	\IEEEcompsocitemizethanks{

		\IEEEcompsocthanksitem This work was supported in part by the National Natural Science Foundation of China under Grants 62271418 and 61871335, and in part by the Natural Science Foundation of Sichuan Province under Grant 2023NSFSC0030.
        (Corresponding author: Heng-Chao Li)

		Yitao Zhao, Nanqing Liu, and Heng-Chao Li are with the School of Information Science and Technology, Southwest Jiaotong University, Chengdu 611756, China (e-mail: ytzhao@my.swjtu.edu.cn; lansing163@163.com; lihengchao78@163.com). Feng Gao is with the School of Information Science and Engineering, Ocean University of China, Qingdao 266100, China. Turgay Celik is with the School of Information Science and Technology, Southwest Jiaotong University, Chengdu 611756, China, also with the School of Electrical and Information Engineering, University of the Witwatersrand, Johannesburg 2000, South Africa, and also with the Faculty of Engineering and Science, University of Agder, 4604 Kristiansand, Norway (e-mail: celikturgay@gmail.com).}
}

\IEEEtitleabstractindextext{
        \vspace{-0.5em}
	\begin{abstract}

        In conventional remote sensing change detection (RS CD) procedures, extensive manual labeling for bi-temporal images is first required to maintain the performance of subsequent fully supervised training. However, pixel-level labeling for CD tasks is very complex and time-consuming. In this paper, we explore a novel self-supervised contrastive framework applicable to the RS CD task, which promotes the model to accurately capture spatial, structural, and semantic information through domain adapter and hierarchical contrastive head. The proposed SSLChange framework accomplishes self-learning only by taking a single-temporal sample and can be flexibly transferred to main-stream CD baselines. With self-supervised contrastive learning, feature representation pre-training can be performed directly based on the original data even without labeling. After a certain amount of labels are subsequently obtained, the pre-trained features will be aligned with the labels for fully supervised fine-tuning. Without introducing any additional data or labels, the performance of downstream baselines will experience a significant enhancement.  Experimental results on 2 entire datasets and 6 diluted datasets show that our proposed SSLChange improves the performance and stability of CD baseline in data-limited situations. The code of SSLChange will be released at \url{https://github.com/MarsZhaoYT/SSLChange}
	\end{abstract}

	}

\maketitle
\IEEEdisplaynontitleabstractindextext
\IEEEpeerreviewmaketitle

\section{Introduction}

\IEEEPARstart{C}{hange} detection (CD) task plays a crucial role in Land Use and Land Cover (LULC), which aims at detecting and highlighting the changed regions in bi-temporal and multi-temporal remote sensing (RS) image sequences \cite{liu2019review}. 
\begin{figure}[!htp]
	\centering
	\includegraphics[width=0.95\linewidth]{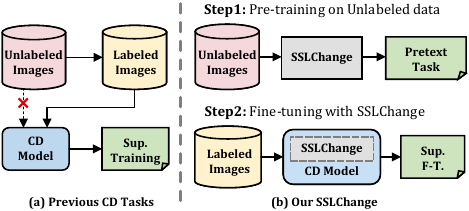}
	\caption{\textbf{Previous CD tasks vs. Our SSLChange.} Illustration of Fine-Tuning method with clipped pre-trained encoder. \textit{F.-T.} represents down-stream fine-tuning.}
	\label{paradiam}
\end{figure}
Benefiting from the rapid growth of aerospace imaging platforms, a variety of RS images with different resolutions and modalities have greatly facilitated breakthroughs in change detection technology. Generally, the change detection framework takes registered RS images of the same location acquired at different times as input and identifies changed objects \cite{cheng2023change}. Considering the wide coverage area of remote sensing images, CD technology automatically and efficiently retrieves changed pixels from large-scale images, greatly reducing the pressure of manual interpretation \cite{shafique2022deep}. Therefore, CD technology is widely applied in urban planning, disaster assessment, environmental monitoring, and other fields \cite{jiang2022survey}.

The combination of CD and computer vision has derived numerous excellent algorithms in RS CD tasks \cite{zheng2022ems, chen2022fccdn, zhang2020feature, chen2021remote, wang2022cbam, li2022transunetcd, mei2021hyperspectral}. The algorithm accuracy and efficiency of CD technology have been significantly improved with the dual support of computing power and data resources, promoting a wide range of applications in the fields of land use planning and street view detection. However, the remarkable successes of such data-driven methods cannot conceal the fatal drawback of severe reliance on large amounts of manual annotations \cite{liu2019review}.
From the perspective of algorithm strategy, the above-mentioned CD methods belong to the applications of fully supervised learning, in which the deep features extracted from the deep models are compared with the manually annotated ground truth, and then the model is guided to optimize by well-designed loss evaluation metrics. Due to the diversity of RS data format and the requirement for prior knowledge of interpretation, only a few datasets with annotated information are accessible for CD tasks \cite{liu2022remote}. Data annotation remains a challenging procedure in the RS community. Some researchers propose several unsupervised CD algorithms to handle the label-limited constraint and obtain promising results \cite{hu2023prbcd, wu2023fully, hu2022total, zhou2023progressive, noh2022unsupervised}. To alleviate the limitation of insufficient annotated data, a feasible solution is to perform data augmentation on the labeled datasets to enlarge the data volume \cite{9852465}. Commonly employed data augmentation methods include random cropping and shifting, rotation and flipping, random scaling, brightness adjustment, and noise addition \cite{shorten2019survey, hao2023review}. Another formulation of data augmentation is to perform random masking or mixing on the input patches, such as Mixup \cite{zhang2017mixup}, Cutout \cite{devries2017improved} and CutMix \cite{yun2019cutmix}. These augmentation methods have sparked remarkable performance in computer vision downstream tasks, but the lack of labeled data remains an obstacle in the RS community.

While fully supervised learning methods have achieved excellent performance, the research interest has gradually shifted to self-supervised learning (SSL) methods. In SSL-based tasks, the model is encouraged to autonomously explore the latent feature from the unlabeled data by constructing pretext tasks. Then the visual representation ability is transferred to downstream tasks. Representative SSL-based methods proposed for vision tasks like \cite{he2020momentum, chen2020simple, grill2020bootstrap, chen2021exploring} enable models to obtain satisfying performance on unlabeled datasets, even surpassing fully supervised methods \cite{chen2021exploring}. Researchers have attempted to introduce SSL-based methods in RS vision tasks \cite{xue2022self, berg2022self, jian2022ss, tao2023tov}.

However, most of the existing SSL-based RS tasks focus on object-level tasks, and there are only few explorations on pixel-level processing such as CD tasks \cite{10126079, calhoun2022self, zhao2022remote, hu2022hypernet, li2023multiform, 10193882, 10188870, wei2021aligning, li2023aligndet}. In the absence of pixel-level annotations, it is difficult for existing SSL-based vision models to be immune to various image representation differences in bi-temporal RS images only relying on global feature vector comparison. Specifically, the limitations of existing SSL-based methods on RS CD tasks are as follows. $1)$ Adaptability of SSL-based frameworks. The mainstream deep learning-based models are not specially designed for pixel-level CD tasks on RS images, ignoring the local spatial features while focusing on global features. In addition, few targeted approaches are able to adaptively handle the nonlinear gray-scale difference caused by the different climate conditions at bi-temporal imaging moments. $2)$ Post-processing procedures. The prediction results of most existing SSL-based CD methods suffer from image noise and uncertain regions. Subsequent post-processing methods like down-stream fine-tuning or threshold segmentation are always utilized to refine the model performance \cite{wu2021unsupervised, luppino2019unsupervised}. However, details of the such post-processing will not be fully revealed. 

Overall, existing CD methods rely on numerous manually labeled data to optimize the model. Without the guidance of labels or only a handful of labels available, the performance of the model will be significantly degraded. Under this circumstance, the ability of the network structure is virtually weakened, and the model has a high probability of over-fitting. \emph{Can we explore the potential of the visual representation in a self-supervised strategy to alleviate the dependence on labels?} Motivated by this, we rethink SSL-based CD procedures. If we can acquire image pairs from different domains through adaptive methods, these inter-domain samples could be identified as natural positive labels. Therefore, SSL pre-training can be firstly performed between the inter-domain samples to get feature representations. After adequate labels are obtained, the pre-trained representations will serve as guidance for down-streamed fully supervised fine-tuning to improve model performance.

In this paper, we propose a self-supervised framework for bi-temporal change detection on RS images via domain adaption. The SSLChange framework is organized in a two-stage approach, containing an inter-domain adaptive encoder and a hierarchical contrastive head successively. The domain adapter adversarially gathers the distance between the original bi-temporal images and eliminates the effects of different imaging conditions. And the hierarchical contrastive head is assigned to increase the similarity between features from unchanged regions. It is worth noting that since SSLChange adopts a SimSiam-like paradigm, the network only receives pseudo positive pairs in latent space, which are generated by the domain adapter from the same single-temporal image. The purpose of such an operation is to improve the robustness to pseudo changes caused by data augmentation, focusing on the structural changes of ground objects at different imaging temporal.

The main contributions can be summarized as follows:

\begin{itemize}

    \item We propose a self-supervised framework named \textbf{SSLChange} for bi-temporal CD tasks on RS images that requires no additional data or labels, which can be conveniently integrated into existing CD baselines. 
    

    \item We develop an \textbf{Domain Adapter} module for bi-temporal SSL-based pretext tasks. In comparison to the original data augmentation, this transformation aligns more closely with real-world scenarios.
    

    \item We design a \textbf{Hierarchical Contrastive Head} including spatial branch and channel branch to extract local and global features to effectively exploit pixel-level semantic information.

\end{itemize}

The rest of this paper is organized as follows. Section \uppercase\expandafter{\romannumeral2} expounds previous work related to this paper. The details of the proposed SSLChange framework are described in Section \uppercase\expandafter{\romannumeral3}. Extensive experimental results are presented in Section \uppercase\expandafter{\romannumeral4}. Finally, the summary and perspectives are drawn in Section 
 \uppercase\expandafter{\romannumeral5}.

\section{Related Work}

\subsection{Domain Adaptation in RS Data}
Inter-domain variance is a common problem in multi-temporal remote sensing tasks \cite{li2021learning, zhang2022hyperspectral}. Since multi-temporal images are acquired with different sensor parameters and external environments, uncertain inter-domain differences are introduced to the data \cite{yang2024multi}. Traditional manual interpretation is based on expert a priori knowledge \cite{li2019deep, sun2021research}. However, traditional statistical models may recognize such inter-domain differences as different patterns. With the increase of hardware computing power, convolutional neural network (CNN) is found to possess a certain degree of invariance to the semantic differences of multi-temporal remote sensing images and the extracted features are semantically consistent \cite{chaman2021truly, crawford2019spatially, mei2021accelerating}. However, when the inter-domain differences are further expanded, such as dual-temporal images acquired in sunny and rainy seasons, the convolutional neural network fails to distinguish the inter-domain variance on the images \cite{zhang2022domain, chen2020geometric, zhang2023cross}. To tackle the inter-domain variance, researchers have proposed the concept of domain adaptation, hoping to mine the mapping between latent spaces among the multi-temporal data. Optimal latent features are expected to have unique correspondence on the original multi-temporal images. Subsequent feature reconstruction is based on the invariant latent features, and the reconstructed multi-temporal features share similar distributions.

The following two solutions to inter-domain variance are generally adopted: distribution metric consistency-based approaches and adversarial-based approaches. Auto-encoder is commonly employed in the former approaches, which performs feature reconstruction on embedding samples and measure the distance between the original samples and the reconstructed features to evaluate the performance \cite{su2019daen}. Zhang et al. propose an unsupervised framework for multi-resolution change detection tasks. By the use of denoising auto-encoder, the inter-domain variance between different images is largely eliminated \cite{zhang2016change}. Sun et al. introduce coupled anto-encoder into bi-temporal HSI Image CD task, coping with the feature differences caused by spectral variation \cite{sun2023intrinsic}. However, the dimension of reconstructed features through auto-encoder is lower than the input samples, which brings errors to the distance metric. Furthermore, auto-encoder is prone to "Mode collapse" during training \cite{10177760, zhang2020deep}. Specifically, instead of learning the distribution of the input data, the model keeps outputting the same results as the input data. In this particular case, the loss of the model maintains a low value, but the performance cannot be satisfying \cite{dong2018review}.

The typical architecture for adversarial-based approaches is the generative adversarial network (GAN) \cite{goodfellow2020generative}. GAN performs feature extraction and reconstruction from random input noise vectors to generate fake samples similar to real ones. Meanwhile, the discriminator receives both real samples from the dataset and fake samples from the generator and strives to judge the authenticity of the input data. The parameters of the GAN are optimized through iterative competition between the generator and the discriminator \cite{creswell2018generative}. Chen et al. propose a CD model with adversarial augmentation to handle the data insufficiency \cite{chen2021adversarial}. Li et al. modify the vanilla GAN to translate optical and SAR images into the same feature domain, then perform CD between the translated heterogeneous data \cite{li2021deep}. However, the vanilla GAN may also suffer from the extreme case of mode collapse during training. Moreover, since the GAN generates objects from random noise vectors, the generated results are inevitably affected by random inputs, which is unacceptable in some pixel-level vision tasks \cite{shi2022latent, ham2020unbalanced}.

Overall, the above work demonstrates the benefit of domain adaptation methods in vision tasks, but the reconstructed features still need to be further refined to rule out wrong generation results. Therefore, it is necessary to construct a robust domain adaptation module to better serve downstream CD tasks.

\subsection{Self-supervised Contrastive Learning in CD task}
Self-supervised representation learning has demonstrated success in NLP, but it still has great potential for visual tasks requiring pixel-level dense prediction. With the inspiration of self-supervision, visual tasks are undergoing a whole new stage of development, of which contrastive self-supervised methods are an essential branch. Some contrastive self-supervised methods such as MoCo\cite{he2020momentum}, SimCLR\cite{chen2020simple}, BYOL\cite{grill2020bootstrap}, SimSiam\cite{chen2021exploring} have enabled models pre-trained in label-free ways to outperform supervised pre-training models on many downstream tasks. In such contrastive methods, images are encoded into positive-negative pairs \cite{he2020momentum, chen2020simple}or double positive pairs\cite{grill2020bootstrap, chen2021exploring}, then the siamese architecture is utilized for feature extraction and comparison.

For the huge amount of data in the RS community, fine-grained labeling is a nearly impossible burden. The boom of contrastive self-supervised methods makes it possible to perform unlabeled pre-training in RS tasks, especially in RS CD tasks. Manas et al. propose a multi-augmentation contrastive self-supervised framework  \cite{manas2021seasonal}, named Seasonal Contrast (SeCo), which takes the effect of seasonal changes on feature representations into consideration to perform change detection between Sentinel-2 satellite images. Jiang et al. propose a two-branch contrastive framework by embedding the CD backbone into a self-supervised paradigm to extract local and global features, which outperforms the supervised baseline \cite{jiang2023self}. Muhtar et al. propose a contrastive mask image distillation framework in teacher–student self-distillation architecture \cite{muhtar2023cmid}.

Although the aforementioned methods highly improve the efficiency and contribution of unlabeled data in RS CD tasks, random data augmentation is still required to construct pseudo-contrastive sample pairs when performing self-supervised pre-training \cite{wang2022self, li2022global, chen2021self, jia2023collaborative}. Uncertainty may be introduced during this process, causing fluctuations in model performance. In this regard, the proposed SSLChange framework takes into account the characteristics of multi-temporal RS CD tasks, introducing domain adaptive into CD tasks to suppress the representation differences between multi-temporal contrastive sample pairs to enhance the adaptability on inter-domain tasks.

\begin{figure*}[!htbp]
	\centering
        \scalebox{1.0}{
	\includegraphics[width=\linewidth]{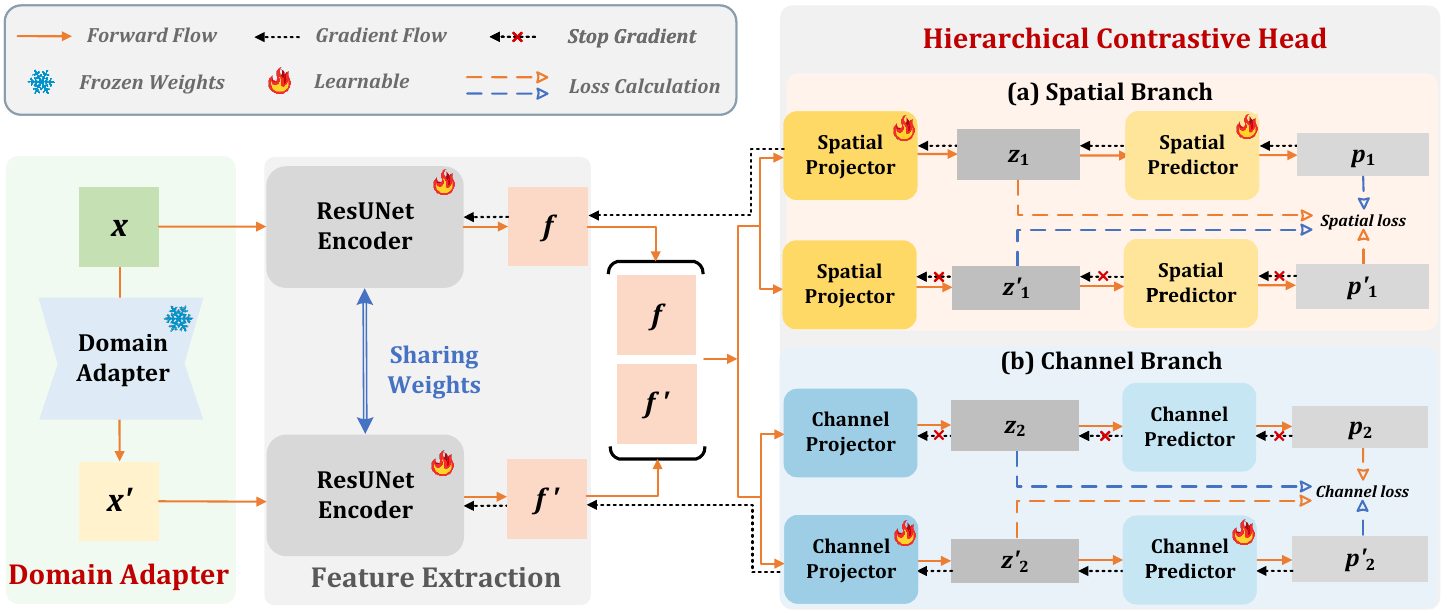}}
	\caption {Workflow of the proposed SSLChange framework for RS CD tasks. The whole SSLChange framework contains two main parts: Domain Adapter and Hierarchical Contrastive Head. First, a transferred view $x^{\prime}$ is obtained through the domain adapter from a single temporal sample $x$. Then the paired views $\left\{x, x^{\prime}\right\}$ is processed by a ResUNet Encoder with sharing weights to get feature set $\left\{f, f^{\prime}\right\}$. The feature set is assigned to two branches: The spatial branch and the channel branch. Each branch consists of a projector and a predictor. The spatial branch captures geometric features by coupled convolutional units, while the channel branch distills the semantic feature by dimension reduction. }
	\label{sslchange}
\end{figure*}

\section{Proposed Method}
In this section, we first elaborately introduce the overview of the proposed SSLChange framework. Then the main components of SSLChange are described sequentially. The specific method of transferring the pre-trained SSLChange framework to downstream RS CD tasks is also explained.
\subsection{Overview}

The overall workflow of SSLChange is illustrated in Figure \ref{sslchange}. Our approach builds upon the distillation-based Self-Supervised Learning (SSL) paradigm \cite{he2020momentum,chen2020simple,chen2021exploring}. However, in contrast to the aforementioned methods, we originally propose two key components applicable to the RS CD task: \textbf{Domain Adapter} and \textbf{Hierarchical Contrastive Head}. The Domain Adapter facilitates the transformation of bi-temporal patches from the CD dataset into the latent space, subsequently reconstructing paired views. Following this, an encoder is used to extract features from paired images. Unless specifically stated, we employ ResUNet as the encoder. Finally, the Hierarchical Contrastive Head processes the paired features, performing feature contrast and similarity calculations within the Siamese branches. Next, we will provide a detailed introduction to these modules.





\subsection{Domain Adapter}
Many current SSL-based approaches \cite{he2020momentum,chen2020simple,chen2021exploring} heavily depend on manual data augmentation to create diverse embedding views, necessitating manual tuning tailored to different pretext tasks. This often entails numerous trials involving various augmentation techniques to identify the optimal combination \cite{chen2020simple}. In the context of Remote Sensing Change Detection (RS CD) tasks, we revisit this process. Given that change detection involves pairs of images captured at distinct time points, influenced by varying imaging conditions such as lighting and weather, we propose a different approach. Leveraging the variations in image style, we train a domain adapter for bidirectional projection. This strategy enables us to generate a pair of naturally aligned positive samples.

To achieve this goal, the most straightforward way is to utilize existing image-to-image translation (I2IT) algorithms \cite{isola2017image,zhu2017unpaired,liu2017unsupervised} to input images. Here, we illustrate using a GAN-based I2IT algorithm as a representative example. Given a pair of images $\left\{x_1, x_2\right\}\in\mathbb{R}^{H*W*C}$, where $x_1$ and $x_2$ share the same geolocation but differ in time phase and image styles, we establish a set of image translation and discriminative networks $(G_1, D_1)$ and $(G_2, D_2)$:

\begin{equation}
	x_1^{\prime}=G_1(x_1), \quad x_2^{\prime}=G_2(x_2)
\end{equation}
where $x_1^{\prime}$ and $x_2^{\prime}$ are the reconstructed samples. The discriminator $D_1$ and $D_2$ need to judge whether the input sample is generated or natural. $G_1$ and $G_2$ are trained adversarially to obtain the ability to project between the two image domains, which is able to generate positive samples. According to the analysis in our previous work \cite{zhao2022comparative}, we choose modified CycleGAN \cite{zhu2017unpaired} with stable performance as the domain adapter.

\begin{figure}[!htbp]
	\centering
	\includegraphics[width=0.63\linewidth]{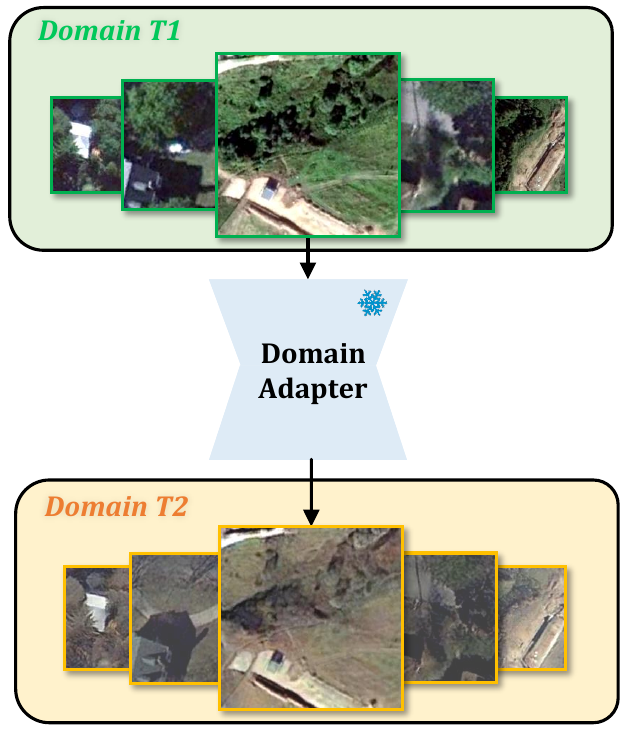}
	\caption{Illustration of the Domain Adapter for bi-temporal transformation on binary CD dataset. The domain adapter is used to project the samples from domain $T_1$ to the opposite domain $T_2$. }
	\label{da}
\end{figure}

As shown in Fig. \ref{da}, we freeze the model parameters of $G_1 / G_2$ as Domain Adapter $\Theta_{ada}$, which is used for the generation of transferred views. A single-temporal patch $x\in\mathbb{R}^{H*W*C}$ is sampled from the first temporal $T_1$ subset of the change detection dataset. Its corresponding transferred view is $x^{\prime}=\Theta_{ada}(x)$. View $x^{\prime}$ generated by $\Theta_{ada}$ differs from the source view $x$ in image style, but shares the same content with view $x$. Therefore, we obtain paired views $\left\{x, x^{\prime}\right\}$ from the generated results, which is naturally suitable for image contrastive learning.  It is worth noting that in the specific implementation of SSLChange, only $G_1$ is employed as the domain adapter. The experimental detail can be found in Section \ref{sec:ablation}.

\subsection{Hierarchical Contrastive Head}
As shown in Fig. \ref{sslchange}, the generated paired views $\left\{x, x^{\prime}\right\}$ will be fed into the encoder for feature extraction and Hierarchical Contrastive Head for further processing. The hierarchical contrastive head contains a spatial branch and a channel branch. Each branch contains a projector and a predictor. 

\subsubsection{\textbf{Spatial and Channel Coding}}
In the spatial branch above, $x$ and $x^{\prime}$ pass through the spatial projector and predictor respectively to get intermediate code $\left\{z, z^{\prime}\right\}$ and deep code $\left\{p, p^{\prime}\right\}$. The specific calculation is as follows:
\begin{equation}
    \left\{\begin{array}{l}
    z_1=\mathbf{Spa\_Proj}(\mathbf{E}(x)), \quad p_1=\mathbf{Spa\_Pred}(z_1) \\
    z_1^{\prime}=\mathbf{Spa\_Proj}(\mathbf{E}(x^{\prime})), \quad p_1^{\prime}=\mathbf{Spa\_Pred}(z_1^{\prime})
    \end{array}\right.
\end{equation}
where \textbf{E} is the ResUNet encoder. Similarly, $x$ and $x^{\prime}$ pass through the channel projector and predictor respectively in the below branch to obtain $\left\{z_2, z_2^{\prime}\right\}$ and $\left\{p_2, p_2^{\prime}\right\}$.

\begin{equation}
	\left\{\begin{array}{l}
		z_2=\mathbf{Cha\_Proj}(\mathbf{E}(x)), \quad p_2=\mathbf{Cha\_Pred}(z_2) \\
		z_2^{\prime}=\mathbf{Cha\_Proj}(\mathbf{E}(x^{\prime})), \quad p_2^{\prime}=\mathbf{Cha\_Pred}(z_2^{\prime})
	\end{array}\right.
\end{equation}

\textbf{Architecture settings.} The specific architecture of the components in the SSLChange framework is as follows. 

\begin{itemize}
	\item \textit{ResUNet Encoder}. The ResUNet Encoder uses the U-Net structure with a ResNet-18 backbone. The size of the output feature is consistent with the input. The features from layers [2, 4, 5, 6, 7] are extracted and unsampled for fusion.
	\item \textit{Spatial Projector $\&$ Predictor}. The Spatial Projector and Predictor share the same structure, both consisting of 2 basic convolutional units with BN and ReLU applied to them. The size of the output feature is consistent with the input.
	\item \textit{Channel Projector $\&$ Predictor}. The Channel Projector consists of a 3-layer MLP with a BN layer. The output layer has no BN or activation layer. The Channel Predictor is a 2-layer MLP. The former layer applies BN. The output of both modules is a 2048-dimensional vector. 
\end{itemize}

\subsubsection{\textbf{Hierarchical Contrastive Learning}}
We employ spatial and channel dimension operations to dig the deep structure and semantic features in RS images, then feature contrast is performed based on the aforementioned features. It is worth noting that we do not introduce any additional labels in this step, but only a hierarchical cross-contrastive mechanism is established between the pipelines in each branch to constrain the network optimization. This prompts the model to decrease its reliance on labels by the self-supervised contrastive mechanism, further exploiting the spatial and semantic representational capabilities of the model.

\begin{table}[!htbp]
	\centering
	\refstepcounter{table}
	\label{pseudo_code}%
	\footnotesize
	\renewcommand{\arraystretch}{1.02}
	\setlength{\tabcolsep}{5pt}
	{
		\begin{tabular}{l}
			\bottomrule
			\textbf{Algorithm 1} SSLChange PseudoCode, PyTorch-like \\
			\hline	\\ [-0.3ex]
				
			\textcolor{codegreen}{\texttt{\# DA: Domain Adapter}} \\ [-0.3ex]
			\textcolor{codegreen}{\texttt{\# E: ResUNet Encoder}} \\ [-0.3ex]
			\textcolor{codegreen}{\texttt{\# Spa\_Proj/Pred: Spatial projector/predictor}} \\ [-0.3ex]
			\textcolor{codegreen}{\texttt{\# Cha\_Proj/Pred: Channel projector/predictor}} \\ [-0.3ex]
			
			\\ [-0.3ex]
			\texttt{\textcolor{codepink}{for} x in \textcolor{codepink}{loader}:} \\ [-0.5ex]
			\qquad \texttt{x' = DA(x)  \textcolor{codegreen}{\# generate transferred view x'}} \\ [-0.5ex]
			\qquad \texttt{f, f' = E(x), E(x')  \textcolor{codegreen}{\# extract features}} \\  [-0.3ex]
			\\ [-0.3ex]
			\qquad \texttt{\textcolor{codegreen}{\# Spatial projections\& predictions}} \\ [-0.3ex]
			\qquad \texttt{z1, z1' = Spa\_Proj(f), Spa\_Proj(f')} \\ [-0.3ex]
			\qquad \texttt{p1, p1' = Spa\_Pred(z1), Spa\_Ped(z1')} \\ [-0.3ex]
			\\  [-0.3ex]
			\qquad \texttt{\textcolor{codegreen}{\# Channel projections\& predictions}} \\ [-0.3ex]
			\qquad \texttt{z2, z2' = Cha\_Proj(f), Cha\_Proj(f')} \\ [-0.3ex]
			\qquad \texttt{p2, p2' = Cha\_Pred(z2), Cha\_Ped(z2')} \\ [-0.3ex]
			\\ [-0.3ex]
			\qquad \texttt{\textcolor{codegreen}{\# calculate losses}} \\ [-0.3ex]
			\qquad \texttt{L\_Spa = D(z1, p1')/2 + D(z1', p1)/2} \\ [-0.3ex]
			\qquad \texttt{L\_Cha = D(z2, p2')/2 + D(z2', p2)/2} \\ [-0.3ex]
			\qquad \texttt{L = (L\_Spa + L\_Cha)/2} \\ [-0.3ex]
			\\ [-0.3ex]
			\qquad \texttt{L.backward() \quad \textcolor{codegreen}{\# back-propagate}} \\ [-0.3ex]
			
			\qquad \texttt{update(E)} \quad \texttt{\textcolor{codegreen}{\# update parameters}} \\ [-0.3ex]
			\qquad \texttt{update(Spa\_Proj, Spa\_Pred)} \\ [-0.3ex]
			\qquad \texttt{update(Cha\_Proj, Cha\_Pred)} \\ [-0.3ex]
			\\ [-0.3ex]
			\texttt{\textcolor{codepink}{def} D(p, z)  \textcolor{codegreen}{\# negative cosine similarity}} \\ [-0.3ex]
			\qquad \texttt{z = z.detach()  \textcolor{codegreen}{\# stop gradient}} \\ [-0.3ex] \\ [-0.3ex]
			\qquad \texttt{p = normalize(p, dim=1)  \textcolor{codegreen}{\# L2-norm on p}} \\ [-0.3ex]
			\qquad \texttt{z = normalize(z, dim=1)  \textcolor{codegreen}{\# L2-norm on z}} \\ [-0.3ex]
			\qquad \texttt{\textcolor{codepink}{return} -(p*z).sum(dim=1).mean()} \\ [-0.3ex]
			
			\\ [-0.3ex]	\toprule
		\end{tabular}
	}
\end{table}

To achieve this, the feature distance between pipelines in each branch needs to be pulled closer. Denoting the two output codes as $z_1$ and $p_1^{\prime}$, we minimize the negative cosine similarity between two codes:
\begin{equation}
	D(z_1, p_1^{\prime}) = - \left \langle \frac{z_1}{\Vert z_1\Vert_{2}},  		 \frac{p_1^{\prime}}{\Vert p_1^{\prime}\Vert_{2}} \right \rangle
\end{equation}

Specifically, we obtain the spatial codes sets ${z_1, p_1}$ and ${z_1^{\prime}, p_1^{\prime}}$ through previous calculations. The calculation for loss in the spatial branch is:

\begin{equation}
	\mathcal{L}_{Spa} = \frac{D(z_1, p_1^{\prime})}{2} + \frac{D(z_1^{\prime}, p_1)}{2}
\end{equation}

Similarly, the loss in the channel branch is calculated by:

\begin{equation}
	\mathcal{L}_{Cha} = \frac{D(z_2, p_2^{\prime})}{2} + \frac{D(z_2^{\prime}, p_2)}{2}
\end{equation}

 The total loss is the linearly weighted sum of the spatial contrastive loss and the channel contrastive loss:

\begin{equation}
	\mathcal{L} = \alpha\cdot\mathcal{L}_{Spa} + (1-\alpha) \cdot \mathcal{L}_{Cha}
\end{equation}
where $\alpha$ represents the weight parameter for the spatial branch.

Through the aforementioned optimization, the model achieves a balance between the spatial coding and semantic coding branches and possesses the capability to extract image structural and semantic features more accurately. It should be noted that, during the optimization process in each branch, the stop-gradient operation is performed on one of the two pipelines, and the gradient will only be back-propagated along the other pipeline. This operation ensures that the model avoids collapsing in the absence of negative samples. The experimental detail can be found in Section \ref{sec:ablation}. 

To facilitate a better comprehension of our approach, we present the pseudo-code of the proposed SSLChange framework in Algorithm~\ref{pseudo_code}.

\subsection{Downstream Fine-tuning}


In this section, we delve into the process of implementing the SSL-based pre-trained model for downstream pixel-level tasks, addressing a crucial and often overlooked detail.

While existing works predominantly focus on applying SSL-based pre-training to object-level or instance-level tasks like image classification or object detection, there is a significant gap in understanding how to adapt this approach to pixel-level tasks, specifically change detection. In the majority of cases, SSL-based pre-training involves tasks at the image level where the pre-trained encoder's output vector conveniently serves as the input to the bottleneck layer during transfer to downstream tasks. However, the scenario becomes more intricate when dealing with pixel-level tasks such as change detection. In this context, the baseline expects an input image with dimensions of $\left(C,H,W\right)$, which does not readily align with the output vector dimension (e.g., 2048) generated by the pre-trained encoder.

We first attempt to directly freeze the pre-trained encoder as a feature extractor and connect it to the CD baseline. However, we find that the outputs of the pre-trained encoder usually contain high-dimensional semantic features, which cannot intuitively reflect the target distribution as the original image. In addition, such rigid migration may lead to gradient vanishing and performance degradation in CD tasks. 

\begin{figure}[ht]
	\centering
	\includegraphics[width=\linewidth]{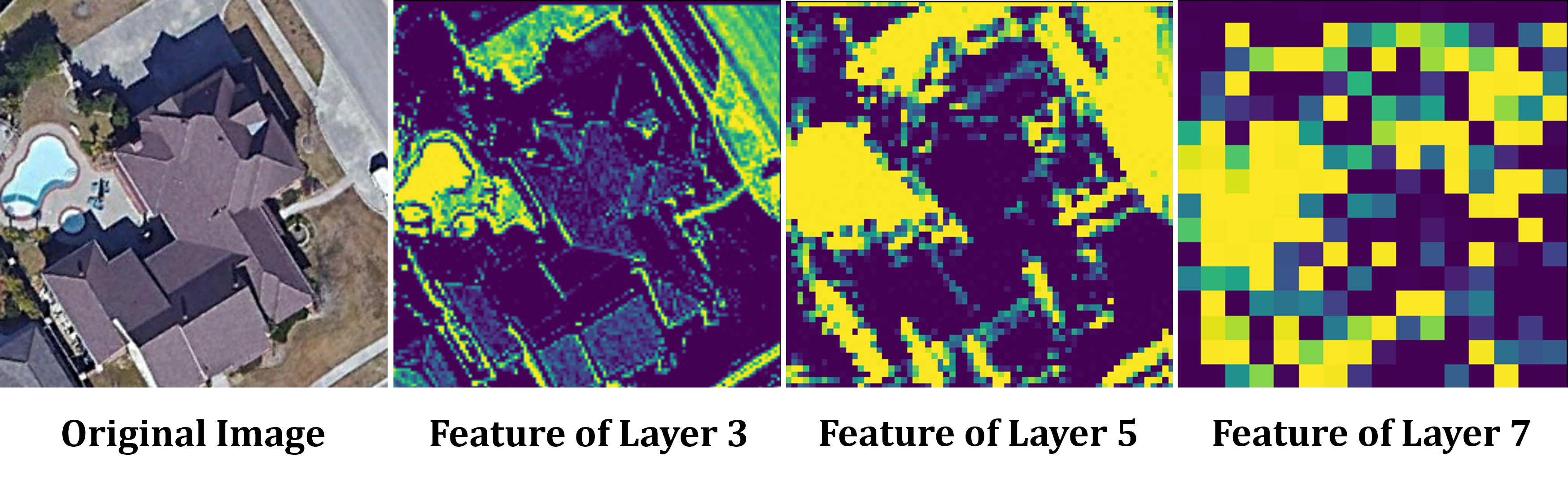}
	\caption{Visualization of pre-trained features from different layers.}
	\label{features}
\end{figure}

For a clear presentation, we visualize the pre-trained features from the frozen encoder in Fig. \ref{features}. Considering that the shallow features contain more structural features such as edges and corners, we clip the pre-trained encoder and only extract the first 3 shallow features of the output. To solve the problem of dimensional unalignment between the output features and the original image, we find that only a lightweight alignment module containing 2 deconvolutional layers is sufficient to maintain the model performance to the utmost degree.

As shown in Fig. \ref{finetuning}, given a pair of images $\left\{T1, T2\right\}$ to be detected, we use the clipped pre-trained encoder to extract shallow features, and restore the feature shape through the alignment module. Then the aligned features are concatenated with the original images in the channel dimension as the embedding of the downstream baseline network. Finally, the change map $\left(CM\right)$ calculated by the baseline network is compared with the ground truth $\left(GT\right)$ to fine-tune and optimize by means of supervision. The optimization method is as follows:

\begin{figure}[!htp]
    \centering
    \includegraphics[width=\linewidth]{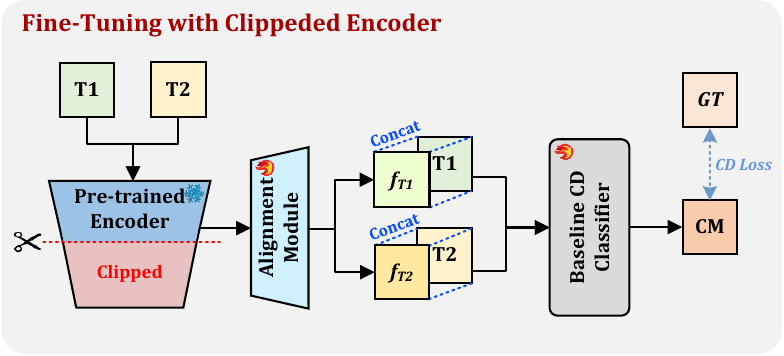}
    \caption{Illustration of Fine-Tuning method with clipped pre-trained encoder. }
    \label{finetuning}
\end{figure}

\begin{equation}
    \mathit{CM} = \mathcal{F}\left(\mathit{Concat}\left(T_{1}, f_{T1} \right), \mathit{Concat}\left(T_{2}, f_{T2} \right)  \right) 
\end{equation}

\begin{equation}
    \mathbf{min} \quad \mathcal{L}\left(\mathit{CM, GT} \right)
\end{equation}

\vspace{-0.1cm}
where \textit{CM} is the change map calculated by baseline, and \textit{GT} is the ground truth. $\mathcal{F}$ is the downstream baseline and $\mathcal{L}$ is the loss function corresponding to the downstream baseline.  In the case of SNUNet \cite{fang2021snunet}, we denote the detailed calculation of the loss function as follows:

\begin{equation}
    \left\{\begin{array}{l}
        \mathcal{L}\left(\mathit{y, \hat{y}} \right) = \mathcal{L}_{WCE} + \mathcal{L}_{Dice} \vspace{1ex}
        \\
        \mathcal{L}_{WCE} = - \omega \cdot \mathit{\hat{y}} \cdot log(\mathit{y}) + (1-\mathit{\hat{y}}) \cdot (log(1-\mathit{y})) \vspace{1ex}
        \\
        \mathcal{L}_{Dice} = 1 - \dfrac{2 \cdot \mathit{\hat{y}} \cdot softmax(\mathit{y})}{\mathit{\hat{y}} + softmax(\mathit{y})}
    \end{array}\right. 
\end{equation}
where $\omega$ is the weight parameter for the positive samples. $y$ and $\hat{y}$ are the change map and ground truth, respectively.

The experimental detail of down-stream fine-tuning can be found in Section \ref{quantitative} and Section \ref{visual}. 

\vspace{0.3cm}
\section{Experiments}
\subsection{Datasets}
In this section, we introduce the datasets used in the experiments. It is worth noting that we want to verify the effectiveness of the proposed SSLChange framework in the extreme case of limited data, so each selected dataset is randomly diluted by the ratio of 10\%, 20\%, and 50\%. Comparative experiments are performed both on 6 diluted datasets and 2 original datasets simultaneously.
\vspace{0.1cm}
\subsubsection{LEVIR-CD Series Datasets}
The LEVIR-CD dataset \cite{chen2020spatial} contains 673 VHR bi-temporal remote sensing image pairs, and the resolution of each image is 0.5m. The original size of the image is 1024$\times$1024 pixels and the entire dataset is split by the ratio of 7:2:1 into train/validation/test sets. To ensure consistent testing standards, the number of test sets was not diluted. The main labeling object is the changes in buildings between two time phases. The specific settings of LEVIR-CD series datasets are shown in Table \ref{levir_settings}.
\vspace{0.1cm}
\subsubsection{CDD Series Datasets}
The CDD dataset \cite{lebedev2018change} contains 16,000 pairs of VHR remote sensing images, with resolutions ranging from 0.03m to 1m. The size of the original image is 256*256 pixels. The provider of the CDD dataset splits the entire dataset into 10000/3000/3000, so we keep this split in the experiments. The main labeling objects are changes in buildings, roads, and vehicles. The specific settings of CDD series datasets are shown in Table \ref{cdd_settings}.

\begin{table}[!htp]
	\centering
	\caption{Settings of LEVIR-CD Series Datasets. \\(The numbers in the table are a pair of images)}
	\label{levir_settings}
	\renewcommand{\arraystretch}{1.4}
	\setlength{\tabcolsep}{8.5pt}
	\footnotesize
	{
		\begin{tabular}{c|c|c|c|c}
			\hline
			\textbf{Subset} & \textbf{Resolution} & \textbf{Train} & \textbf{Validation} & \textbf{Test}
			\\ \hline
			LEVIR-CD-10\%  & \multirow{4}{*}{0.5 m/pixel} & 45 & 13 & \multirow{4}{*}{64}
			\\ \cline{1-1} \cline{3-4}
			LEVIR-CD-20\%  & 	& 89 & 26 & 
			\\ \cline{1-1} \cline{3-4}
			LEVIR-CD-50\%  & 	& 223  & 64 & 
			\\ \cline{1-1} \cline{3-4}
			LEVIR-CD 		& 	& 445 & 128 & 
			\\ \hline 
		\end{tabular}
	}
\end{table}

\begin{table}[!htp]
	\centering
	\caption{Settings of CDD Series Datasets. \\(The numbers in the table are a pair of images)}
	\label{cdd_settings}
	\renewcommand{\arraystretch}{1.4}
	\setlength{\tabcolsep}{8pt}
	\footnotesize
	{
		\begin{tabular}{c|c|c|c|c}
			\hline
			\textbf{Subset} & \textbf{Resolution} & \textbf{Train} & \textbf{Validation} & \textbf{Test}
			\\ \hline
			CDD-10\%  & \multirow{4}{*}{0.03m-1 m/pixel} & 1000 & 300 & \multirow{4}{*}{3000}
			\\ \cline{1-1} \cline{3-4}
			CDD-20\%  & 	& 2000 & 600 & 
			\\ \cline{1-1} \cline{3-4}
			CDD-50\%  & 	& 5000  & 1500 & 
			\\ \cline{1-1} \cline{3-4}
			CDD 	  & 	& 10000 & 3000 & 
			\\ \hline 
		\end{tabular}
	}
\end{table}

Note that we use the entire unlabeled LEVIR-CD and CDD datasets during the SSLChange pre-training. Then the diluted datasets are used for downstream fine-tuning to evaluate the effectiveness of the pre-trained encoder from the SSLChange framework under limited data.

\subsection{Evaluation Metrics}
In evaluating the performances of the SSLChange framework, we select four evaluation metrics: Precision, Recall, F1-Score, and IoU. The specific calculation method for metrics is as follows:
\begin{small}
\begin{equation}
    \begin{array}{l}
    \mathit{Precision} = \dfrac{TP}{TP + FP} \vspace{1.4ex} \\ 
    \mathit{Recall} = \dfrac{TP}{TP + FN}  \vspace{1.4ex} \\ 
    \mathit{F1} = 2 \times \dfrac{Precision \times Recall}{Precision + Recall}  
    \vspace{1.4ex} \\ 
    \mathit{IoU} = \dfrac{TP}{TP + FN + FP}
    \end{array}
\end{equation} 
\end{small}
where, \textit{TP}, \textit{FP}, and \textit{FN} are components in the confusion matrix, representing true positive, false positive, and false negative, respectively. 

\vspace{-0.1cm}
\subsection{Implementation Details}
\subsubsection{Selected Downstream RS CD Baselines}
We select five open-source RS CD baselines to evaluate the effectiveness of the proposed SSLChange framework transferred to them. 

\begin{itemize}
	\item FC-EF \cite{daudt2018fully}: Fully convolutional network with a single-branch U-Net architecture. Bi-temporal images are fused early before entering the network. Skip connection is used in the upsampling and downsampling stages.
	
	\item FC-Siam-conc \cite{daudt2018fully}: Dual-branch siamese architecture with full convolutional network. Each branch receives an image of a single phase and performs feature concatenation fusion and long-range skip connection during the upsampling stage.
	
	\item FC-Siam-diff \cite{daudt2018fully}: Dual-branch siamese architecture with full convolutional network. The difference map is calculated during the downsampling stage. Feature concatenation fusion and long-range skip connection are performed during the upsampling stage.
	
	\item SNUNet-CD \cite{fang2021snunet}: Dual-branch U-Net++ architecture. Hierarchical features of the input images are connected densely, and the ECAM attention mechanism is performed to focus on the features.
	
	\item USSFCNet \cite{lei2023ultralightweight}: A lightweight architecture design is adopted. Multi-scale decoupled convolution and spatial-spectral feature cooperation strategy is introduced to extract richer features.
	
\end{itemize}

\subsubsection{Model Training and Testing}
The training and inference processes are implemented on an NVIDIA RTX 3090 GPU with 24GB of memory. The SSLChange framework is trained for 100 epochs with a batch size of 8. The parameters are optimized by SGD optimizer with momentum of 0.9 and weight decay of 0.0001. The initial learning rate is set to 0.001 with cosine decay. In the downstream fine-tuning, the selected baselines are optimized by the AdamW optimizer with a momentum of 0.999 and weight decay of 0.01. Considering the limited data volume in downstream tasks, the batch size for fine-tuning is set to 4. Furthermore, we discard all random operations in the augmentation during downstream fine-tuning to eliminate the influence of randomness, and only format transformation functions are retained.

\subsection{Experimental Results}
\subsubsection{Quantitative Analysis}
\label{quantitative}
In this part, we perform comparative experiments on the selected CD baselines on 2 entire datasets and 6 diluted datasets. Specifically, we evaluate the improvement of the proposed SSLChange framework over existing CNN-based CD baselines, especially the performance and stability in data-limited situations.

\begin{table*}[!htbp]
	\centering
	\caption{Comparison results on datasets with different sampling ratios from CDD dataset. The green subscripts in the table represent the performance gain of the baselines with the SSLChange framework applied compared to the original baselines.}
	\label{tab:cdd}
		\renewcommand{\arraystretch}{1.65}
		\footnotesize
		\setlength{\tabcolsep}{4.2pt}
		\scalebox{0.83}{
			\begin{tabular}{c c |cccc| cccc| cccc| cccc}
				\Xhline{1pt}
				\multicolumn{1}{c}{\multirow{2}{*}{Baseline}} 
				& \multicolumn{1}{c|}{\multirow{2}{*}{+SSLChange}} 
				& \multicolumn{4}{c|}{\textbf{CDD-10\%}} & \multicolumn{4}{c|}{\textbf{CDD-20\%}} 
				& \multicolumn{4}{c|}{\textbf{CDD-50\%}} & \multicolumn{4}{c}{\textbf{CDD}} 
				\\
                \multicolumn{2}{c|}{} 
				& \multicolumn{1}{c}{Precision} & \multicolumn{1}{c}{Recall} 
				& \multicolumn{1}{c}{\textbf{F1}} & \multicolumn{1}{c|}{\textbf{IoU}} 
				& \multicolumn{1}{c}{Precision} & \multicolumn{1}{c}{Recall} 
				& \multicolumn{1}{c}{\textbf{F1}} & \multicolumn{1}{c|}{\textbf{IoU}}
				& \multicolumn{1}{c}{Precision} & \multicolumn{1}{c}{Recall} 
				& \multicolumn{1}{c}{\textbf{F1}} & \multicolumn{1}{c|}{\textbf{IoU}}
				& \multicolumn{1}{c}{Precision} & \multicolumn{1}{c}{Recall} 
				& \multicolumn{1}{c}{\textbf{F1}} & \multicolumn{1}{c}{\textbf{IoU}}
				\\
				\hline
				\multirow{3}{*}{FC-EF \cite{daudt2018fully}}
				& \ding{55} 
				& 69.62 & 64.91 & 67.18 &50.58
				& 89.51 & 62.72 & 73.76 &58.43
				& 90.52 & 77.04 & 83.24 &71.29
				& 95.23 & 79.39 & 86.59 &76.35
				\\ 
				\cline{2-18}            
				& \ding{51}
				& 77.10 & 60.37 
				& 67.72 
				& 51.19
				& 81.99 & 70.40 
				& 75.75
				& 60.97
				& 86.03 & 76.52 
				& 81.01
				& 68.07
				& 91.43 & 83.83 
				& 87.47  
				& 77.72
				\\ \cline{2-18}  
                & $\Delta$
                & & & \textcolor{darkgreen}{+0.54} & \textcolor{darkgreen}{+0.61}
                
                & & & \textcolor{darkgreen}{+1.99} & \textcolor{darkgreen}{+2.54}
                
                & & & -2.23 & -3.22
                
                & & & \textcolor{darkgreen}{+0.88} & \textcolor{darkgreen}{+1.37} 
                \\
                
				\hline
				\multirow{3}{*}{FC-Siam-conc \cite{daudt2018fully}}
				& \ding{55} 
				& 81.63 & 38.51 & 52.34 &35.44
				& 83.23 & 44.06 & 57.62 &40.46
				& 88.96 & 48.19 & 62.51 &45.47
				& 94.62 & 34.21 & 50.25 &33.56
				\\
				\cline{2-18}            
				& \ding{51}
				& 79.16 & 40.08
				& 53.22
				& 36.26
				& 79.82 & 46.15 
				& 58.49
				& 41.33
				& 85.47 & 52.27 
				& 64.87
				& 48.01
				& 90.01 & 51.50
				& 65.51  
				& 48.71
				\\
				\cline{2-18}  
				& $\Delta$
				& & & \textcolor{darkgreen}{+0.88} & \textcolor{darkgreen}{+0.82}
				
				& & & \textcolor{darkgreen}{+0.87}  & \textcolor{darkgreen}{+0.87}
				
				& & & \textcolor{darkgreen}{+2.36}  & \textcolor{darkgreen}{+2.54}
				
				& & & \textcolor{darkgreen}{+15.36} & \textcolor{darkgreen}{+15.15} 
				\\
				
				\hline
				\multirow{3}{*}{FC-Siam-diff \cite{daudt2018fully}}
				& \ding{55} 
				& 85.98 & 30.81 & 45.36 &29.33
				& 88.72 & 37.96 & 53.17 &36.21
				& 92.15 & 37.63 & 53.44 &36.46
				& 93.26 & 36.35 & 52.31 &35.42
				\\
				\cline{2-18}            
				& \ding{51}
				& 83.05 & 35.32
				& 49.57
				& 32.95
				& 86.88 & 39.21 
				& 54.03
				& 37.02
				& 88.59 & 43.09
				& 57.98
				& 40.82
				& 90.91 & 47.47
				& 62.37
				& 45.32
				\\
				\cline{2-18}  
				& $\Delta$
				& & & \textcolor{darkgreen}{+4.21} & \textcolor{darkgreen}{+3.62}
				
				& & & \textcolor{darkgreen}{+0.86} & \textcolor{darkgreen}{+0.81}
				
				& & & \textcolor{darkgreen}{+4.54}  & \textcolor{darkgreen}{+4.36}
				
				& & & \textcolor{darkgreen}{+10.06}   & \textcolor{darkgreen}{+9.90} 
				\\
				
				\hline
				\multirow{3}{*}{SNUNet \cite{fang2021snunet}}
				& \ding{55} 
				& 71.69 & 69.80 & 70.73 &54.71
				& 91.83 & 60.14 & 72.68 &57.09
				& 91.32 & 74.46 & 82.03 &69.54
				& 92.51 & 83.99 & 88.05 &78.64
				\\
				\cline{2-18}            
				& \ding{51}
				& 96.76 & 65.57
				& 74.69
				& 59.60
				& 89.37 & 73.16 
				& 80.46
				& 67.30
				& 90.97 & 79.98 
				& 85.12
				& 74.10
				& 90.10 & 79.12 
				& 84.26
				& 72.80
				\\
				\cline{2-18}  
				& $\Delta$
				& & & \textcolor{darkgreen}{+3.96}  & \textcolor{darkgreen}{+4.89}
				
				& & & \textcolor{darkgreen}{+7.78}  & \textcolor{darkgreen}{+10.21}
				
				& & & \textcolor{darkgreen}{+3.09}  & \textcolor{darkgreen}{+4.56}
				
				& & & -3.79  & -5.84
				\\
				
				\hline
				\multirow{3}{*}{USSFCNet \cite{lei2023ultralightweight}}
				& \ding{55} 
				& 84.94 & 57.34 & 68.46 &52.05
				& 87.73 & 72.70 & 79.51 &65.99
				& 89.58 & 80.99 & 85.07 &74.01
				& 90.79 & 91.40 & 91.10 &83.65
				\\
				\cline{2-18}            
				& \ding{51}
				& 81.01 & 64.56
				& 71.86
				& 56.07
				& 83.56 & 77.21
				& 80.26
				& 67.03
				& 90.41 & 84.44
				& 87.32
				& 77.49
				& 92.82 & 90.03
				& 91.40
				& 84.17 
				\\
				\cline{2-18}  
				& $\Delta$
				& & & \textcolor{darkgreen}{+3.40} & \textcolor{darkgreen}{+4.02}
				
				& & & \textcolor{darkgreen}{+0.75} & \textcolor{darkgreen}{+1.04}
				
				& & & \textcolor{darkgreen}{+2.25} & \textcolor{darkgreen}{+3.48}
				
				& & & \textcolor{darkgreen}{+0.30} & \textcolor{darkgreen}{+0.52}
				\\
				\Xhline{1pt}
			\end{tabular}
		}
	\end{table*}

\begin{table*}[!htbp]
	\centering
	\caption{Comparison results on datasets with different sampling ratios from the LEVIR-CD dataset. The green subscripts in the table represent the performance gain of the baselines with the SSLChange framework applied compared to the original baselines.}
	\label{tab:levir}
		\renewcommand{\arraystretch}{1.65}
		\footnotesize
		\setlength{\tabcolsep}{4.2pt}
		\scalebox{0.83}{
			\begin{tabular}{c c |cccc| cccc| cccc| cccc}
				\Xhline{1pt}
				\multicolumn{1}{c}{\multirow{2}{*}{Baseline}} 
				& \multicolumn{1}{c|}{\multirow{2}{*}{+SSLChange}} 
				& \multicolumn{4}{c|}{\textbf{LEVIR-10\%}} & \multicolumn{4}{c|}{\textbf{LEVIR-20\%}} 
				& \multicolumn{4}{c|}{\textbf{LEVIR-50\%}} & \multicolumn{4}{c}{\textbf{LEVIR}} 
				\\
                \multicolumn{2}{c|}{} 
				& \multicolumn{1}{c}{Precision} & \multicolumn{1}{c}{Recall} 
				& \multicolumn{1}{c}{\textbf{F1}} & \multicolumn{1}{c|}{\textbf{IoU}} 
				& \multicolumn{1}{c}{Precision} & \multicolumn{1}{c}{Recall} 
				& \multicolumn{1}{c}{\textbf{F1}} & \multicolumn{1}{c|}{\textbf{IoU}}
				& \multicolumn{1}{c}{Precision} & \multicolumn{1}{c}{Recall} 
				& \multicolumn{1}{c}{\textbf{F1}} & \multicolumn{1}{c|}{\textbf{IoU}}
				& \multicolumn{1}{c}{Precision} & \multicolumn{1}{c}{Recall} 
				& \multicolumn{1}{c}{\textbf{F1}} & \multicolumn{1}{c}{\textbf{IoU}}
				\\
				\hline
				\multirow{3}{*}{FC-EF \cite{daudt2018fully}}
				& \ding{55} 
				& 68.58 & 58.96 & 63.40 &46.41
				& 78.01 & 73.42 & 75.65 &60.83
				& 82.87 & 74.13 & 78.26 &64.28
				& 87.11 & 74.01 & 80.03 &66.71
				\\ 
				\cline{2-18}            
				& \ding{51}
				& 67.11 & 61.66 
				& 64.27
				& 47.35
				& 76.87 & 71.73 
				& 74.21
				& 59.01
				& 81.15 & 76.49 
				& 78.75
				& 64.95
				& 83.20 & 79.77 
				& 81.45 
				& 68.70
				\\ \cline{2-18} 
                    & $\Delta$
                    & & & \textcolor{darkgreen}{+0.87} & \textcolor{darkgreen}{+0.94}
                    & & & -1.24 & -1.82
                    & & & \textcolor{darkgreen}{+0.49} & \textcolor{darkgreen}{+0.67}
                    & & & \textcolor{darkgreen}{+1.42} & \textcolor{darkgreen}{+1.99} 
                    \\
				\hline
				\multirow{3}{*}{FC-Siam-conc \cite{daudt2018fully}}
				& \ding{55} 
				& 76.76 & 58.46 & 66.68 &50.01
				& 80.41 & 65.96 & 72.47 &56.83
				& 83.71 & 71.93 & 77.37 &63.09
				& 84.03 & 72.68 & 77.94 &63.86
				\\
				\cline{2-18}            
				& \ding{51}
				& 68.99 & 68.07
				& 68.53
				& 52.12
				& 78.21 & 67.12 
				& 73.85
				& 58.54
				& 82.72 & 77.01 
				& 79.76
				& 66.34
				& 83.53 & 77.62
				& 80.47
				& 67.32
				\\ \cline{2-18} 
                    & $\Delta$
                    & & & \textcolor{darkgreen}{+1.85}  & \textcolor{darkgreen}{+2.11}
                    & & & \textcolor{darkgreen}{+1.38}  & \textcolor{darkgreen}{+1.71}
                    & & & \textcolor{darkgreen}{+2.39}  & \textcolor{darkgreen}{+3.25}
                    & & & \textcolor{darkgreen}{+2.53}  & \textcolor{darkgreen}{+3.46}
                    \\
				
				\hline
				\multirow{3}{*}{FC-Siam-diff \cite{daudt2018fully}}
				& \ding{55} 
				& 76.64 & 28.50 & 41.55 &26.22
				& 83.21 & 61.02 & 70.41 &54.33
				& 84.23 & 64.33 & 72.95 &57.42
				& 84.54 & 72.22 & 77.89 &63.79
				\\
				\cline{2-18}            
				& \ding{51}
				& 74.61 & 54.25
				& 62.83
				& 45.80
				& 81.34 & 63.18 
				& 71.12
				& 55.19
				& 85.52 & 65.64
				& 74.27
				& 59.07
				& 84.88 & 73.34 
				& 78.69
				& 64.86
				\\ \cline{2-18} 
                    & $\Delta$
                    & & & \textcolor{darkgreen}{+21.31} & \textcolor{darkgreen}{+19.58}
                    & & & \textcolor{darkgreen}{+0.71}  & \textcolor{darkgreen}{+0.86}
                    & & & \textcolor{darkgreen}{+1.32}  & \textcolor{darkgreen}{+1.65}
                    & & & \textcolor{darkgreen}{+0.80}  & \textcolor{darkgreen}{+1.07} 
                    \\
				
				\hline
				\multirow{3}{*}{SNUNet \cite{fang2021snunet}}
				& \ding{55} 
				& 65.79 & 58.96 & 62.19 &45.12
				& 68.70 & 59.79 & 63.94 &46.99
				& 83.78 & 64.52 & 72.90 &57.35
				& 78.89 & 67.71 & 72.88 &57.32
				\\
				\cline{2-18}            
				& \ding{51}
				& 64.93 & 62.94
				& 63.92
				& 46.97
				& 70.97 & 61.11 
				& 65.68
				& 48.89
				& 82.46 & 67.91 
				& 75.18
				& 60.23
				& 84.76 & 76.19 
				& 80.25
				& 67.01
				\\ \cline{2-18} 
                    & $\Delta$
                    & & & \textcolor{darkgreen}{+1.73} & \textcolor{darkgreen}{+1.85}
                    & & & \textcolor{darkgreen}{+1.74}  & \textcolor{darkgreen}{+1.90}
                    & & & \textcolor{darkgreen}{+2.28}  & \textcolor{darkgreen}{+2.88}
                    & & & \textcolor{darkgreen}{+7.40}  & \textcolor{darkgreen}{+9.69} 
                    \\
				
				\hline
				\multirow{3}{*}{USSFCNet \cite{lei2023ultralightweight}}
				& \ding{55} 
				& 69.94 & 69.42 & 66.97 &53.46
				& 76.75 & 75.56 & 76.15 &61.49
				& 82.76 & 80.07 & 81.39 &68.62
				& 82.01 & 85.54 & 83.74 &72.02
				\\
				\cline{2-18}            
				& \ding{51}
				& 72.60 & 70.54
				& 71.56
				& 55.71
				& 78.91 & 76.89
				& 77.89
				& 63.78
				& 83.51 & 80.09 
				& 81.76
				& 69.15
				& 84.25 & 83.36
				& 83.81
				& 72.13
				\\ \cline{2-18} 
                    & $\Delta$
                    & & & \textcolor{darkgreen}{+4.59} & \textcolor{darkgreen}{+2.25}
                    & & & \textcolor{darkgreen}{+1.74}  & \textcolor{darkgreen}{+2.29}
                    & & & \textcolor{darkgreen}{+0.37}  & \textcolor{darkgreen}{+0.53}
                    & & & \textcolor{darkgreen}{+0.07}  & \textcolor{darkgreen}{+0.11} 
                    \\
				\Xhline{1pt}
			\end{tabular}
            }
\end{table*}

\noindent \textbf{CDD Series Datasets:} 
The specific results of comparative experiments on the CDD series datasets are shown in Table \ref{tab:cdd}. The results reflect that the proposed SSLChange framework provides a large gain for the existing RS CD baselines on the CDD series datasets. The green subscripts in the table represent the performance gain of the baselines with the SSLChange framework applied compared to the original baselines. On the CDD series datasets, SSLChange showed a positive impact on most of the baselines, with the largest gains in the main evaluation metrics F1 and IoU reaching 15.36\% and 15.15\%, respectively. In terms of precision and recall metrics, SSLChange also showed a great improvement, which shows that the SSLChange framework helps the baselines more accurately and comprehensively detect the real change regions between the bi-temporal images. From the horizontal comparison, it can be found that when the datasets are diluted to 10\% or 20\%, the performance of the original baselines shows a large degradation. During the training, we also found that after removing the random augmentation in the original baseline, the performance of the baselines would become oscillatory, and the results of several training sessions under the same configuration showed drastic fluctuations. While the application of the SSLChange framework can help the baselines return to a relatively stable performance. We notice that there exist 2 outliers in the result table, respectively in FC-EF (on CDD-50\% dataset) and SNUNet-CD (CDD dataset). After analysis, we consider that the fine-tuning strategy applied is to freeze the parameters of the clipped pre-trained encoder for feature extraction, and then an upsampling alignment module is added to the baseline. This operation increases the number of parameters and complexity of the baselines. In addition, the CDD dataset has a relatively large amount of data, so it fails to show advantages under limited training epochs.

\begin{table*}[!htb]
	\centering
	\caption{Visual comparison of baselines and our proposed SSLChange method on CDD Series datasets. }
	\label{visua_cdd}       
	\newcolumntype{?}{!{\vrule width 2pt}}
	\scalebox{0.67}{
		\renewcommand{\arraystretch}{1.3}
		\begin{tabular}{|m{1.8cm}|m{1.8cm}<{\centering}:m{1.8cm}<{\centering}:m{1.8cm}<{\centering}|m{1.8cm}<{\centering}:m{1.8cm}<{\centering}|m{1.8cm}<{\centering}:m{1.8cm}<{\centering}|m{1.8cm}<{\centering}:m{1.8cm}<{\centering}|m{1.8cm}<{\centering}:m{1.8cm}<{\centering}|}			
			\hline
			\multicolumn{1}{|c|}{\multirow{2}{*}{\textbf{Method}}} & 
			\multicolumn{3}{c|}{\textbf{Testing Samples}} &
			\multicolumn{2}{c|}{\textbf{CDD-10$\%$}} & 
			\multicolumn{2}{c|}{\textbf{CDD-20$\%$}} &
			\multicolumn{2}{c|}{\textbf{CDD-50$\%$}} & 
			\multicolumn{2}{c|}{\textbf{CDD}} \\
			\cline{2-12}
			& T1 & T2 & GT & Sup & SSLChange & Sup & SSLChange & Sup & SSLChange & Sup & SSLChange 
			\\
			\hline \hline
			
			\makecell[c]{
				\begin{minipage}[c]{0.7\linewidth}
					\centering
					FC-EF \cite{daudt2018fully}
				\end{minipage}
			} &
			\makecell[c]{
				\begin{minipage}[b]{0.1\textwidth}
					\centering
					{\includegraphics[width=1.05\textwidth]{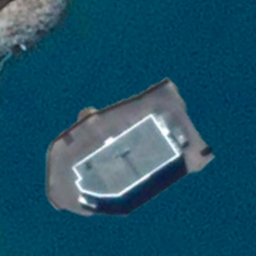}}
				\end{minipage}
			} & 
			\makecell[c]{
				\begin{minipage}[b]{0.1\textwidth}
					\centering
					{\includegraphics[width=\textwidth]{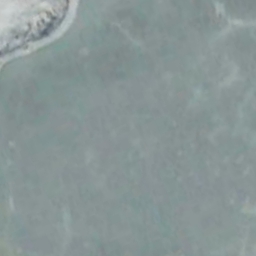}}
				\end{minipage}
			} &
			\makecell[c]{
				\begin{minipage}[b]{0.1\textwidth}
					\centering
					{\includegraphics[width=\textwidth]{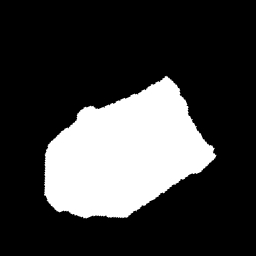}}
				\end{minipage}
			} & 
			\makecell[c]{
				\begin{minipage}[b]{0.1\textwidth}
					\centering
					{\includegraphics[width=\textwidth]{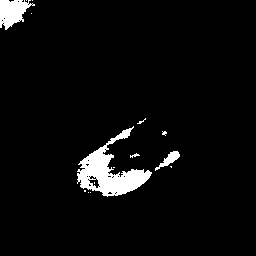}}
				\end{minipage}
			} &
			\makecell[c]{
				\begin{minipage}[b]{0.1\textwidth}
					\centering
					{\includegraphics[width=\textwidth]{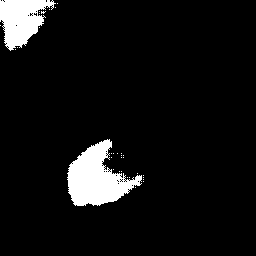}}
				\end{minipage}
			} & 
			\makecell[c]{
				\begin{minipage}[b]{0.1\textwidth}
					\centering
					{\includegraphics[width=\textwidth]{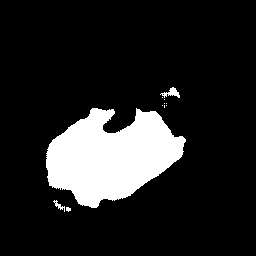}}
				\end{minipage}
			} & 
			\makecell[c]{
				\begin{minipage}[b]{0.1\textwidth}
					\centering
					{\includegraphics[width=\textwidth]{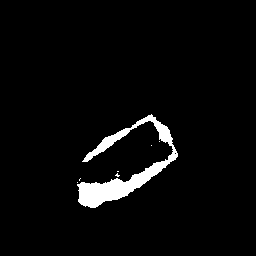}}
				\end{minipage}
			} &
			\makecell[c]{
				\begin{minipage}[b]{0.1\textwidth}
					\centering
					{\includegraphics[width=\textwidth]{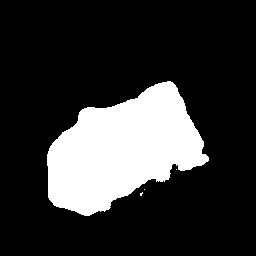}}
				\end{minipage}
			} &
			\makecell[c]{
				\begin{minipage}[b]{0.1\textwidth}
					\centering
					{\includegraphics[width=\textwidth]{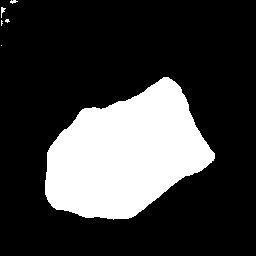}}
				\end{minipage}
			} &
			\makecell[c]{
				\begin{minipage}[b]{0.1\textwidth}
					\centering
					{\includegraphics[width=\textwidth]{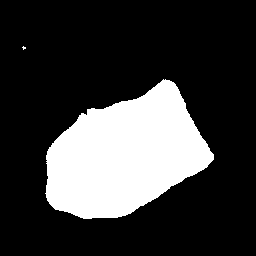}}
				\end{minipage}
			} & 
			\makecell[c]{
				\begin{minipage}[b]{0.1\textwidth}
					\centering
					{\includegraphics[width=\textwidth]{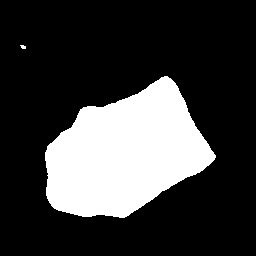}}
			\end{minipage}} 
			\\ \hline 
			
			
			\makecell[c]{
				\begin{minipage}[c]{0.7\linewidth}
					\centering
					FC-Siam-conc \cite{daudt2018fully}
				\end{minipage}
			} &
			\makecell[c]{
				\begin{minipage}[b]{0.1\textwidth}
					\centering
					{\includegraphics[width=\textwidth]{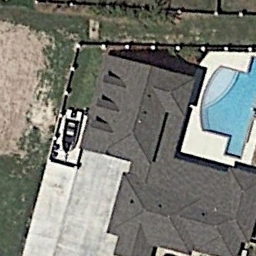}}
				\end{minipage}
			} & 
			\makecell[c]{
				\begin{minipage}[b]{0.1\textwidth}
					\centering
					{\includegraphics[width=\textwidth]{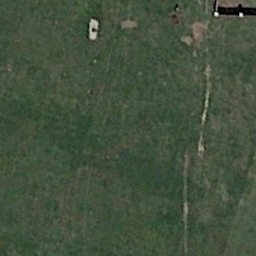}}
				\end{minipage}
			} &
			\makecell[c]{
				\begin{minipage}[b]{0.1\textwidth}
					\centering
					{\includegraphics[width=\textwidth]{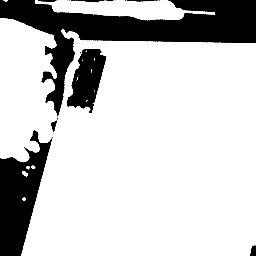}}
				\end{minipage}
			} & 
			\makecell[c]{
				\begin{minipage}[b]{0.1\textwidth}
					\centering
					{\includegraphics[width=\textwidth]{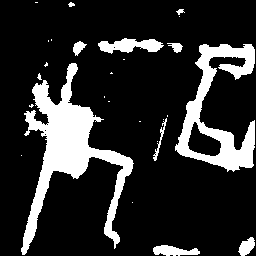}}
				\end{minipage}
			} &
			\makecell[c]{
				\begin{minipage}[b]{0.1\textwidth}
					\centering
					{\includegraphics[width=\textwidth]{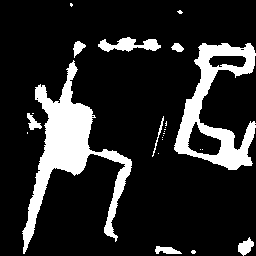}}
				\end{minipage}
			} & 
			\makecell[c]{
				\begin{minipage}[b]{0.1\textwidth}
					\centering
					{\includegraphics[width=\textwidth]{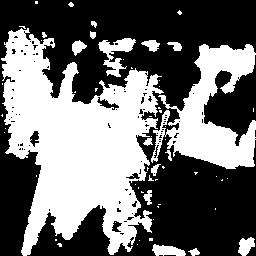}}
				\end{minipage}
			} & 
			\makecell[c]{
				\begin{minipage}[b]{0.1\textwidth}
					\centering
					{\includegraphics[width=\textwidth]{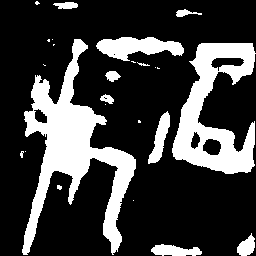}}
				\end{minipage}
			} &
			\makecell[c]{
				\begin{minipage}[b]{0.1\textwidth}
					\centering
					{\includegraphics[width=\textwidth]{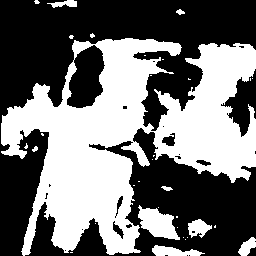}}
				\end{minipage}
			} &
			\makecell[c]{
				\begin{minipage}[b]{0.1\textwidth}
					\centering
					{\includegraphics[width=\textwidth]{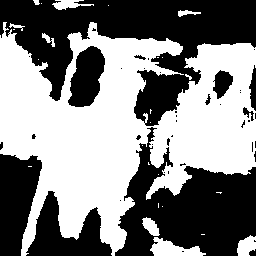}}
				\end{minipage}
			} &
			\makecell[c]{
				\begin{minipage}[b]{0.1\textwidth}
					\centering
					{\includegraphics[width=\textwidth]{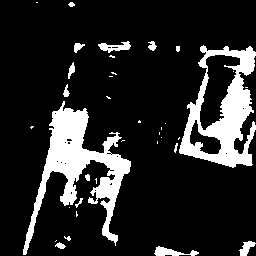}}
				\end{minipage}
			} & 
			\makecell[c]{
				\begin{minipage}[b]{0.1\textwidth}
					\centering
					{\includegraphics[width=\textwidth]{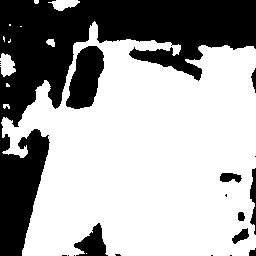}}
			\end{minipage}} 
			\\ \hline
			
			
			\makecell[c]{
				\begin{minipage}[c]{0.7\linewidth}
					\centering
					FC-Siam-diff \cite{daudt2018fully}
				\end{minipage}
			} &
			\makecell[c]{
				\begin{minipage}[b]{0.1\textwidth}
					\centering
					{\includegraphics[width=\textwidth]{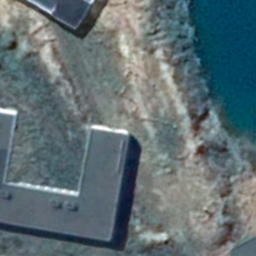}}
				\end{minipage}
			} & 
			\makecell[c]{
				\begin{minipage}[b]{0.1\textwidth}
					\centering
					{\includegraphics[width=\textwidth]{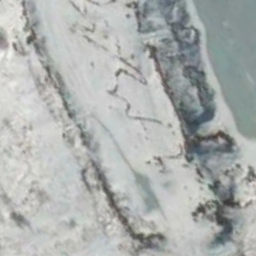}}
				\end{minipage}
			} &
			\makecell[c]{
				\begin{minipage}[b]{0.1\textwidth}
					\centering
					{\includegraphics[width=\textwidth]{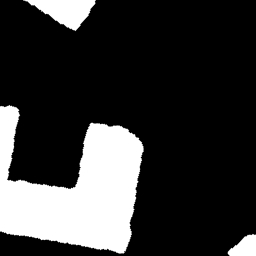}}
				\end{minipage}
			} & 
			\makecell[c]{
				\begin{minipage}[b]{0.1\textwidth}
					\centering
					{\includegraphics[width=\textwidth]{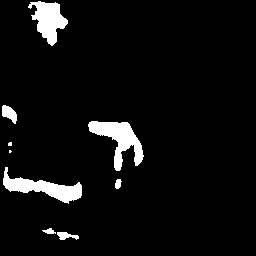}}
				\end{minipage}
			} &
			\makecell[c]{
				\begin{minipage}[b]{0.1\textwidth}
					\centering
					{\includegraphics[width=\textwidth]{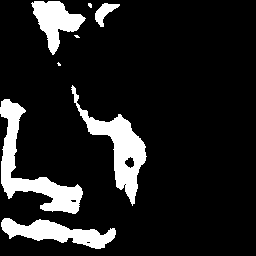}}
				\end{minipage}
			} & 
			\makecell[c]{
				\begin{minipage}[b]{0.1\textwidth}
					\centering
					{\includegraphics[width=\textwidth]{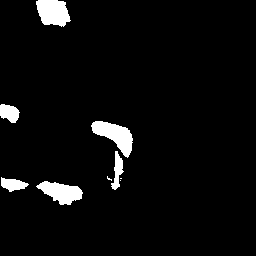}}
				\end{minipage}
			} & 
			\makecell[c]{
				\begin{minipage}[b]{0.1\textwidth}
					\centering
					{\includegraphics[width=\textwidth]{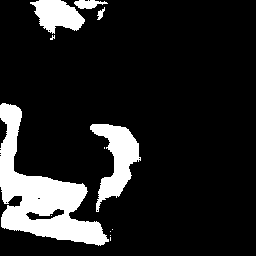}}
				\end{minipage}
			} &
			\makecell[c]{
				\begin{minipage}[b]{0.1\textwidth}
					\centering
					{\includegraphics[width=\textwidth]{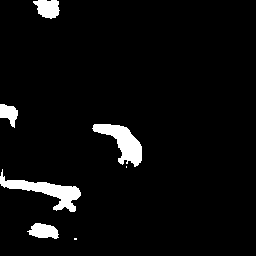}}
				\end{minipage}
			} &
			\makecell[c]{
				\begin{minipage}[b]{0.1\textwidth}
					\centering
					{\includegraphics[width=\textwidth]{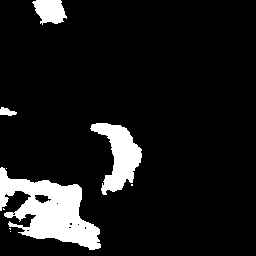}}
				\end{minipage}
			} &
			\makecell[c]{
				\begin{minipage}[b]{0.1\textwidth}
					\centering
					{\includegraphics[width=\textwidth]{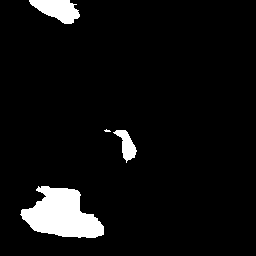}}
				\end{minipage}
			} & 
			\makecell[c]{
				\begin{minipage}[b]{0.1\textwidth}
					\centering
					{\includegraphics[width=\textwidth]{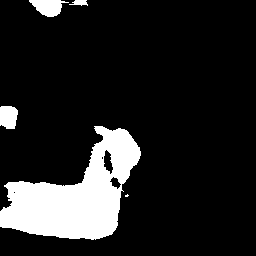}}
			\end{minipage}} 
			\\ \hline
			
			
			\makecell[c]{
				\begin{minipage}[c]{0.7\linewidth}
					\centering
					SNUNet \cite{fang2021snunet}
				\end{minipage}
			} &
			\makecell[c]{
				\begin{minipage}[b]{0.1\textwidth}
					\centering
					{\includegraphics[width=\textwidth]{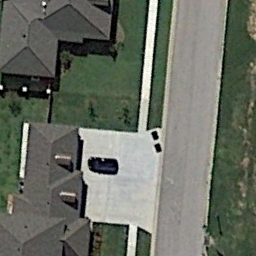}}
				\end{minipage}
			} & 
			\makecell[c]{
				\begin{minipage}[b]{0.1\textwidth}
					\centering
					{\includegraphics[width=\textwidth]{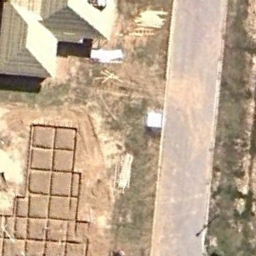}}
				\end{minipage}
			} &
			\makecell[c]{
				\begin{minipage}[b]{0.1\textwidth}
					\centering
					{\includegraphics[width=\textwidth]{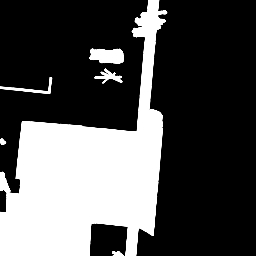}}
				\end{minipage}
			} & 
			\makecell[c]{
				\begin{minipage}[b]{0.1\textwidth}
					\centering
					{\includegraphics[width=\textwidth]{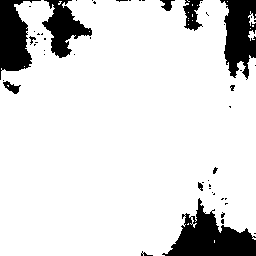}}
				\end{minipage}
			} &
			\makecell[c]{
				\begin{minipage}[b]{0.1\textwidth}
					\centering
					{\includegraphics[width=\textwidth]{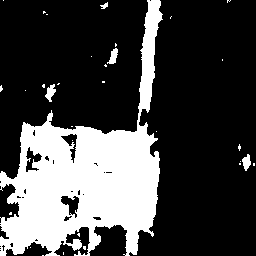}}
				\end{minipage}
			} & 
			\makecell[c]{
				\begin{minipage}[b]{0.1\textwidth}
					\centering
					{\includegraphics[width=\textwidth]{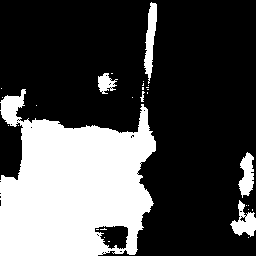}}
				\end{minipage}
			} & 
			\makecell[c]{
				\begin{minipage}[b]{0.1\textwidth}
					\centering
					{\includegraphics[width=\textwidth]{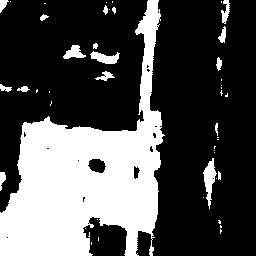}}
				\end{minipage}
			} &
			\makecell[c]{
				\begin{minipage}[b]{0.1\textwidth}
					\centering
					{\includegraphics[width=\textwidth]{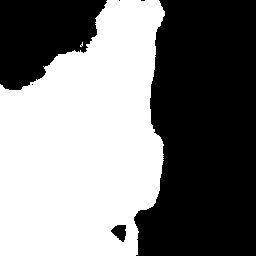}}
				\end{minipage}
			} &
			\makecell[c]{
				\begin{minipage}[b]{0.1\textwidth}
					\centering
					{\includegraphics[width=\textwidth]{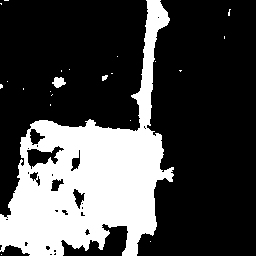}}
				\end{minipage}
			} &
			\makecell[c]{
				\begin{minipage}[b]{0.1\textwidth}
					\centering
					{\includegraphics[width=\textwidth]{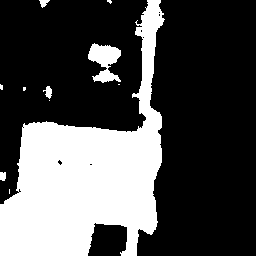}}
				\end{minipage}
			} & 
			\makecell[c]{
				\begin{minipage}[b]{0.1\textwidth}
					\centering
					{\includegraphics[width=\textwidth]{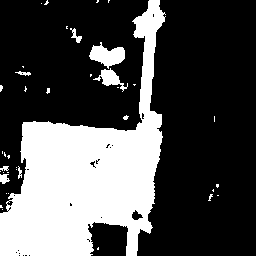}}
			\end{minipage}} 
			\\ \hline
			
			
			\makecell[c]{
				\begin{minipage}[c]{0.7\linewidth}
					\centering
					USSFCNet \cite{lei2023ultralightweight}
				\end{minipage}
			} &
			\makecell[c]{
				\begin{minipage}[b]{0.1\textwidth}
					\centering
					{\includegraphics[width=\textwidth]{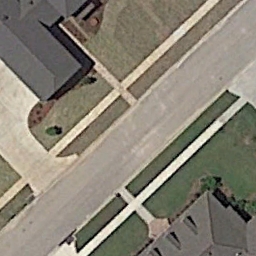}}
				\end{minipage}
			} & 
			\makecell[c]{
				\begin{minipage}[b]{0.1\textwidth}
					\centering
					{\includegraphics[width=\textwidth]{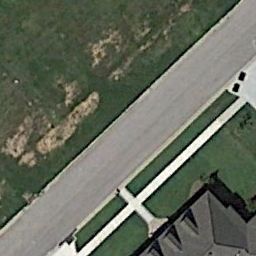}}
				\end{minipage}
			} &
			\makecell[c]{
				\begin{minipage}[b]{0.1\textwidth}
					\centering
					{\includegraphics[width=\textwidth]{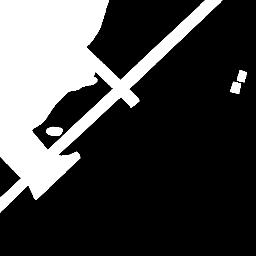}}
				\end{minipage}
			} & 
			\makecell[c]{
				\begin{minipage}[b]{0.1\textwidth}
					\centering
					{\includegraphics[width=\textwidth]{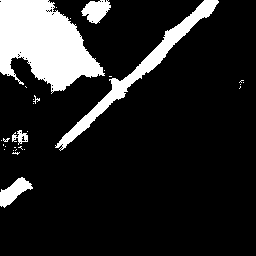}}
				\end{minipage}
			} &
			\makecell[c]{
				\begin{minipage}[b]{0.1\textwidth}
					\centering
					{\includegraphics[width=\textwidth]{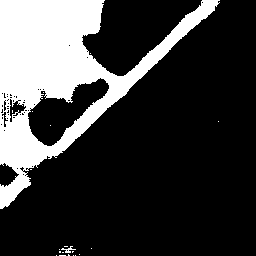}}
				\end{minipage}
			} & 
			\makecell[c]{
				\begin{minipage}[b]{0.1\textwidth}
					\centering
					{\includegraphics[width=\textwidth]{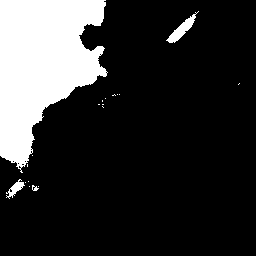}}
				\end{minipage}
			} & 
			\makecell[c]{
				\begin{minipage}[b]{0.1\textwidth}
					\centering
					{\includegraphics[width=\textwidth]{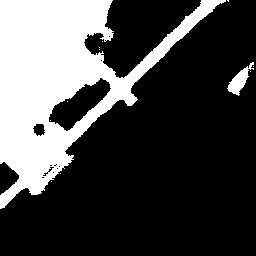}}
				\end{minipage}
			} &
			\makecell[c]{
				\begin{minipage}[b]{0.1\textwidth}
					\centering
					{\includegraphics[width=\textwidth]{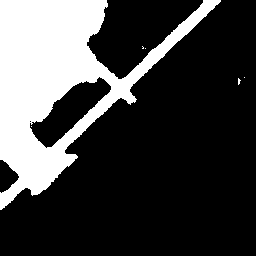}}
				\end{minipage}
			} &
			\makecell[c]{
				\begin{minipage}[b]{0.1\textwidth}
					\centering
					{\includegraphics[width=\textwidth]{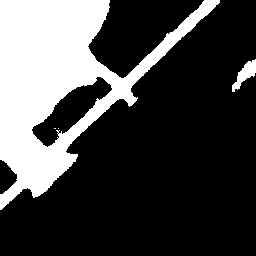}}
				\end{minipage}
			} &
			\makecell[c]{
				\begin{minipage}[b]{0.1\textwidth}
					\centering
					{\includegraphics[width=\textwidth]{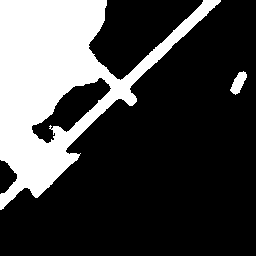}}
				\end{minipage}
			} & 
			\makecell[c]{
				\begin{minipage}[b]{0.1\textwidth}
					\centering
					{\includegraphics[width=\textwidth]{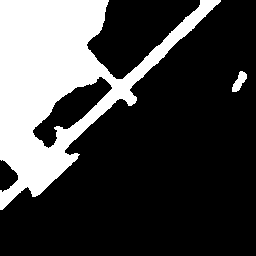}}
			\end{minipage}} 
			\\ \hline
	\end{tabular}}
\end{table*}

\begin{table*}[!htb]
	\centering
	\caption{Visual comparison of baselines and our proposed SSLChange method on LEVIR-CD Series datasets. }
	\label{visual_levir}       
	\newcolumntype{?}{!{\vrule width 2pt}}
	\scalebox{0.67}{
		\renewcommand{\arraystretch}{1.3}
		\begin{tabular}{|m{1.8cm}|m{1.8cm}<{\centering}:m{1.8cm}<{\centering}:m{1.8cm}<{\centering}|m{1.8cm}<{\centering}:m{1.8cm}<{\centering}|m{1.8cm}<{\centering}:m{1.8cm}<{\centering}|m{1.8cm}<{\centering}:m{1.8cm}<{\centering}|m{1.8cm}<{\centering}:m{1.8cm}<{\centering}|}
			
				
				\hline
				\multicolumn{1}{|c|}{\multirow{2}{*}{\textbf{Method}}} & 
				\multicolumn{3}{c|}{\textbf{Testing Samples}} &
				\multicolumn{2}{c|}{\textbf{LEVIR-10$\%$}} & 
				\multicolumn{2}{c|}{\textbf{LEVIR-20$\%$}} &
				\multicolumn{2}{c|}{\textbf{LEVIR-50$\%$}} & 
				\multicolumn{2}{c|}{\textbf{LEVIR}} \\
				\cline{2-12}
				& T1 & T2 & GT & Sup & SSLChange & Sup & SSLChange & Sup & SSLChange & Sup & SSLChange 
				\\
				\hline \hline
				
				\makecell[c]{
					\begin{minipage}[c]{0.7\linewidth}
						\centering
						FC-EF \cite{daudt2018fully}
					\end{minipage}
				} &
				\makecell[c]{
					\begin{minipage}[b]{0.1\textwidth}
						\centering
						{\includegraphics[width=1.05\textwidth]{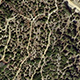}}
					\end{minipage}
				} & 
				\makecell[c]{
					\begin{minipage}[b]{0.1\textwidth}
						\centering
						{\includegraphics[width=\textwidth]{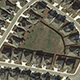}}
					\end{minipage}
				} &
				\makecell[c]{
					\begin{minipage}[b]{0.1\textwidth}
						\centering
						{\includegraphics[width=\textwidth]{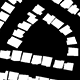}}
					\end{minipage}
				} & 
				\makecell[c]{
					\begin{minipage}[b]{0.1\textwidth}
						\centering
						{\includegraphics[width=\textwidth]{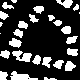}}
					\end{minipage}
				} &
				\makecell[c]{
					\begin{minipage}[b]{0.1\textwidth}
						\centering
						{\includegraphics[width=\textwidth]{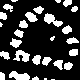}}
					\end{minipage}
				} & 
				\makecell[c]{
					\begin{minipage}[b]{0.1\textwidth}
						\centering
						{\includegraphics[width=\textwidth]{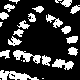}}
					\end{minipage}
				} & 
				\makecell[c]{
					\begin{minipage}[b]{0.1\textwidth}
						\centering
						{\includegraphics[width=\textwidth]{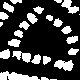}}
					\end{minipage}
				} &
				\makecell[c]{
					\begin{minipage}[b]{0.1\textwidth}
						\centering
						{\includegraphics[width=\textwidth]{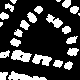}}
					\end{minipage}
				} &
				\makecell[c]{
					\begin{minipage}[b]{0.1\textwidth}
						\centering
						{\includegraphics[width=\textwidth]{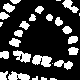}}
					\end{minipage}
				} &
				\makecell[c]{
					\begin{minipage}[b]{0.1\textwidth}
						\centering
						{\includegraphics[width=\textwidth]{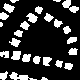}}
					\end{minipage}
				} & 
				\makecell[c]{
					\begin{minipage}[b]{0.1\textwidth}
						\centering
						{\includegraphics[width=\textwidth]{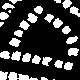}}
				\end{minipage}} 
				\\ \hline 
				
				
				\makecell[c]{
					\begin{minipage}[c]{0.7\linewidth}
						\centering
						FC-Siam-conc \cite{daudt2018fully}
					\end{minipage}
				} &
				\makecell[c]{
					\begin{minipage}[b]{0.1\textwidth}
						\centering
						{\includegraphics[width=\textwidth]{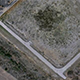}}
					\end{minipage}
				} & 
				\makecell[c]{
					\begin{minipage}[b]{0.1\textwidth}
						\centering
						{\includegraphics[width=\textwidth]{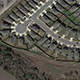}}
					\end{minipage}
				} &
				\makecell[c]{
					\begin{minipage}[b]{0.1\textwidth}
						\centering
						{\includegraphics[width=\textwidth]{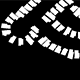}}
					\end{minipage}
				} & 
				\makecell[c]{
					\begin{minipage}[b]{0.1\textwidth}
						\centering
						{\includegraphics[width=\textwidth]{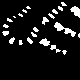}}
					\end{minipage}
				} &
				\makecell[c]{
					\begin{minipage}[b]{0.1\textwidth}
						\centering
						{\includegraphics[width=\textwidth]{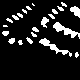}}
					\end{minipage}
				} & 
				\makecell[c]{
					\begin{minipage}[b]{0.1\textwidth}
						\centering
						{\includegraphics[width=\textwidth]{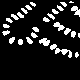}}
					\end{minipage}
				} & 
				\makecell[c]{
					\begin{minipage}[b]{0.1\textwidth}
						\centering
						{\includegraphics[width=\textwidth]{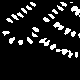}}
					\end{minipage}
				} &
				\makecell[c]{
					\begin{minipage}[b]{0.1\textwidth}
						\centering
						{\includegraphics[width=\textwidth]{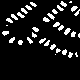}}
					\end{minipage}
				} &
				\makecell[c]{
					\begin{minipage}[b]{0.1\textwidth}
						\centering
						{\includegraphics[width=\textwidth]{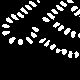}}
					\end{minipage}
				} &
				\makecell[c]{
					\begin{minipage}[b]{0.1\textwidth}
						\centering
						{\includegraphics[width=\textwidth]{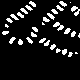}}
					\end{minipage}
				} & 
				\makecell[c]{
					\begin{minipage}[b]{0.1\textwidth}
						\centering
						{\includegraphics[width=\textwidth]{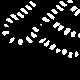}}
				\end{minipage}} 
				\\ \hline
				
				
				\makecell[c]{
					\begin{minipage}[c]{0.7\linewidth}
						\centering
						FC-Siam-diff \cite{daudt2018fully}
					\end{minipage}
				} &
				\makecell[c]{
					\begin{minipage}[b]{0.1\textwidth}
						\centering
						{\includegraphics[width=\textwidth]{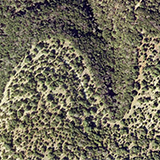}}
					\end{minipage}
				} & 
				\makecell[c]{
					\begin{minipage}[b]{0.1\textwidth}
						\centering
						{\includegraphics[width=\textwidth]{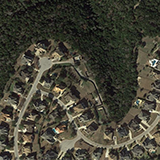}}
					\end{minipage}
				} &
				\makecell[c]{
					\begin{minipage}[b]{0.1\textwidth}
						\centering
						{\includegraphics[width=\textwidth]{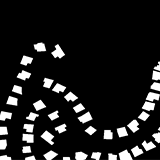}}
					\end{minipage}
				} & 
				\makecell[c]{
					\begin{minipage}[b]{0.1\textwidth}
						\centering
						{\includegraphics[width=\textwidth]{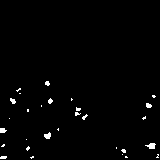}}
					\end{minipage}
				} &
				\makecell[c]{
					\begin{minipage}[b]{0.1\textwidth}
						\centering
						{\includegraphics[width=\textwidth]{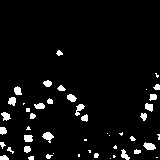}}
					\end{minipage}
				} & 
				\makecell[c]{
					\begin{minipage}[b]{0.1\textwidth}
						\centering
						{\includegraphics[width=\textwidth]{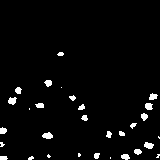}}
					\end{minipage}
				} & 
				\makecell[c]{
					\begin{minipage}[b]{0.1\textwidth}
						\centering
						{\includegraphics[width=\textwidth]{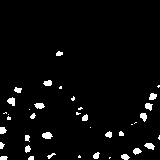}}
					\end{minipage}
				} &
				\makecell[c]{
					\begin{minipage}[b]{0.1\textwidth}
						\centering
						{\includegraphics[width=\textwidth]{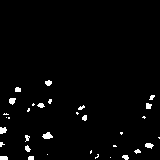}}
					\end{minipage}
				} &
				\makecell[c]{
					\begin{minipage}[b]{0.1\textwidth}
						\centering
						{\includegraphics[width=\textwidth]{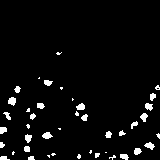}}
					\end{minipage}
				} &
				\makecell[c]{
					\begin{minipage}[b]{0.1\textwidth}
						\centering
						{\includegraphics[width=\textwidth]{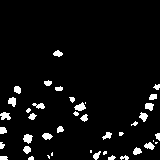}}
					\end{minipage}
				} & 
				\makecell[c]{
					\begin{minipage}[b]{0.1\textwidth}
						\centering
						{\includegraphics[width=\textwidth]{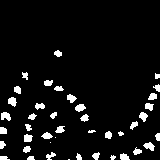}}
				\end{minipage}} 
				\\ \hline

				
				\makecell[c]{
					\begin{minipage}[c]{0.7\linewidth}
						\centering
						SNUNet \cite{fang2021snunet}
					\end{minipage}
				} &
				\makecell[c]{
					\begin{minipage}[b]{0.1\textwidth}
						\centering
						{\includegraphics[width=\textwidth]{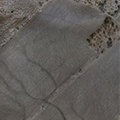}}
					\end{minipage}
				} & 
				\makecell[c]{
					\begin{minipage}[b]{0.1\textwidth}
						\centering
						{\includegraphics[width=\textwidth]{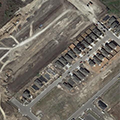}}
					\end{minipage}
				} &
				\makecell[c]{
					\begin{minipage}[b]{0.1\textwidth}
						\centering
						{\includegraphics[width=\textwidth]{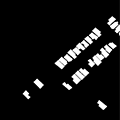}}
					\end{minipage}
				} & 
				\makecell[c]{
					\begin{minipage}[b]{0.1\textwidth}
						\centering
						{\includegraphics[width=\textwidth]{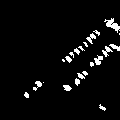}}
					\end{minipage}
				} &
				\makecell[c]{
					\begin{minipage}[b]{0.1\textwidth}
						\centering
						{\includegraphics[width=\textwidth]{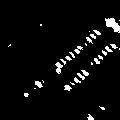}}
					\end{minipage}
				} & 
				\makecell[c]{
					\begin{minipage}[b]{0.1\textwidth}
						\centering
						{\includegraphics[width=\textwidth]{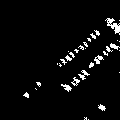}}
					\end{minipage}
				} & 
				\makecell[c]{
					\begin{minipage}[b]{0.1\textwidth}
						\centering
						{\includegraphics[width=\textwidth]{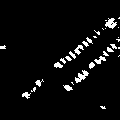}}
					\end{minipage}
				} &
				\makecell[c]{
					\begin{minipage}[b]{0.1\textwidth}
						\centering
						{\includegraphics[width=\textwidth]{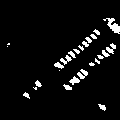}}
					\end{minipage}
				} &
				\makecell[c]{
					\begin{minipage}[b]{0.1\textwidth}
						\centering
						{\includegraphics[width=\textwidth]{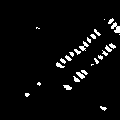}}
					\end{minipage}
				} &
				\makecell[c]{
					\begin{minipage}[b]{0.1\textwidth}
						\centering
						{\includegraphics[width=\textwidth]{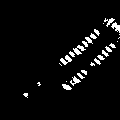}}
					\end{minipage}
				} & 
				\makecell[c]{
					\begin{minipage}[b]{0.1\textwidth}
						\centering
						{\includegraphics[width=\textwidth]{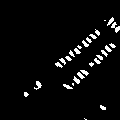}}
				\end{minipage}} 
				\\ \hline
				
				
				\makecell[c]{
					\begin{minipage}[c]{0.7\linewidth}
						\centering
						USSFCNet \cite{lei2023ultralightweight}
					\end{minipage}
				} &
				\makecell[c]{
					\begin{minipage}[b]{0.1\textwidth}
						\centering
						{\includegraphics[width=\textwidth]{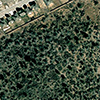}}
					\end{minipage}
				} & 
				\makecell[c]{
					\begin{minipage}[b]{0.1\textwidth}
						\centering
						{\includegraphics[width=\textwidth]{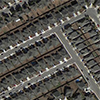}}
					\end{minipage}
				} &
				\makecell[c]{
					\begin{minipage}[b]{0.1\textwidth}
						\centering
						{\includegraphics[width=\textwidth]{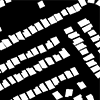}}
					\end{minipage}
				} & 
				\makecell[c]{
					\begin{minipage}[b]{0.1\textwidth}
						\centering
						{\includegraphics[width=\textwidth]{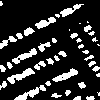}}
					\end{minipage}
				} &
				\makecell[c]{
					\begin{minipage}[b]{0.1\textwidth}
						\centering
						{\includegraphics[width=\textwidth]{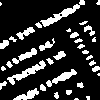}}
					\end{minipage}
				} & 
				\makecell[c]{
					\begin{minipage}[b]{0.1\textwidth}
						\centering
						{\includegraphics[width=\textwidth]{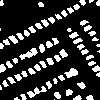}}
					\end{minipage}
				} & 
				\makecell[c]{
					\begin{minipage}[b]{0.1\textwidth}
						\centering
						{\includegraphics[width=\textwidth]{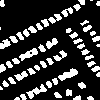}}
					\end{minipage}
				} &
				\makecell[c]{
					\begin{minipage}[b]{0.1\textwidth}
						\centering
						{\includegraphics[width=\textwidth]{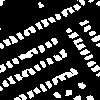}}
					\end{minipage}
				} &
				\makecell[c]{
					\begin{minipage}[b]{0.1\textwidth}
						\centering
						{\includegraphics[width=\textwidth]{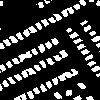}}
					\end{minipage}
				} &
				\makecell[c]{
					\begin{minipage}[b]{0.1\textwidth}
						\centering
						{\includegraphics[width=\textwidth]{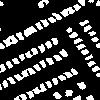}}
					\end{minipage}
				} & 
				\makecell[c]{
					\begin{minipage}[b]{0.1\textwidth}
						\centering
						{\includegraphics[width=\textwidth]{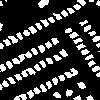}}
				\end{minipage}} 
				\\ \hline
			\end{tabular}
		}
	\end{table*}

\begin{figure*}[!htp]
	\centering
	\includegraphics[width=0.9\linewidth]{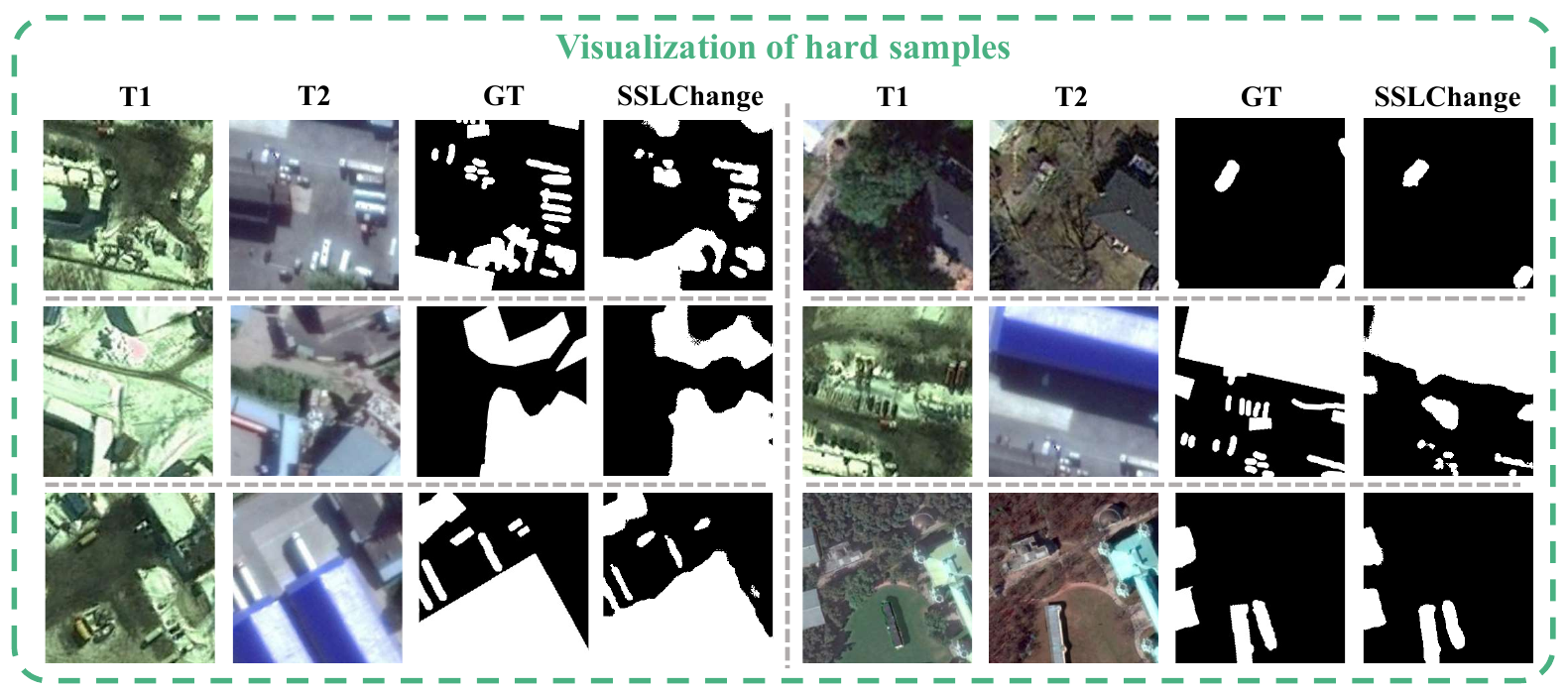}
	\caption{ Performance of the proposed SSLChange in hard samples with complex scenarios.}
	\label{fig:complex}
\end{figure*}

\noindent \textbf{LEVIR-CD Series Datasets:} 
The specific results of comparative experiments on LEVIR-CD series datasets are shown in Table \ref{tab:levir}. The proposed SSLChange framework also shows an improvement effect on the LEVIR-CD series datasets. It is worth noting that the data volume in the LEVIR-CD series datasets is smaller than that of the CDD series datasets. Therefore, the data limitation in the LEVIR-CD series datasets is more serious. In this case, some of the selected baselines also show unstable and over-fitting. When the amount of data is at 50\% or the full ratio, the SSLChange framework can provide satisfied gains for most baselines. While in the extreme cases where the dilution ratio decreases to 10\% and 20\%, the SSLChange framework is able to help the performance of the baselines return to a relatively stable and homogeneous level. The maximum improvement of the F1 and IoU metrics reaches 21.31\% and 19.58\%, which proves the performance advantage of the SSLChange framework in extreme cases with limited data. In addition, we observe an outlier in FC-EF (on the LEVIR-CD-20\% dataset). After a comprehensive analysis, we conclude that FC-EF, as a lightweight network with a simple structure, is more prone to over-fitting in extreme cases with limited data. By comparison, we find that in the LEVIR-CD-20\% dataset, FC-EF presents a falsely high index. The improvement of the SSLChange framework under the small amount of data pursues stability and homogeneity, which is lower than the individual metric of FC-EF. 

On both CDD Series datasets and LEVIR-CD Series datasets, we observe that in some cases, the precision of the proposed SSLChange framework is slightly lower than the baselines. But for the remaining metrics, SSLChange outperforms significantly the baselines. We believe that the reason is that with limited training samples, the baseline is usually overfitted, leading to falsely high precision. Aligned with the SSLChange framework, the baselines exhibit a more generalized recognition capability, allowing the model to achieve higher recall metric at the acceptable expense of accuracy. As a consequence, more regions of real change will be correctly recognized. 

In addition, we find a gap in the trend of increase and decrease in precision and recall between the selected baselines on different datasets. Overall, precision and recall of almost all baselines change asynchronously on the CDD Series datasets. While on the LEVIR-CD Series datasets, precision and recall of SNUNet and USSFCNet increase synchronously. Such performance variation is related to the utilization of the attention mechanisms. SNUNet and USSFCNet employ ECAM and SSFC attentions, respectively. This helps the models to focus on local details in high-resolution images, which is reflected as an increase in precision metric. Moreover, benefiting from the constraint from the SSLChange on the global feature, the recall metrics on LEVIR-CD Series datasets increase simultaneously. However, the samples in the CDD Series dataset are generally of low to medium resolution. The attention mechanism is limited by the image quality to accurately capture local information, leading to a decrease in precision metric. Extensive experimental results show that the SSLChange framework motivates the model to maintain global feature extraction regardless of the quality of the dataset. In realistic scenarios, recall metric is usually given higher priority than precision to ensure that as many changed regions as possible are accurately detected. In addition, considering that F1 is a more comprehensive metric than precision and recall, our proposed SSLChange still obtains the highest performance on F1 despite the decrease in precision.

\subsubsection{Visualization Analysis}
\label{visual}
In order to show the improvement effect of the SSLChange framework more intuitively, we perform visualizations on 2 entire datasets and 6 diluted datasets. The visualization results of the CDD and the LEVIR-CD series datasets are listed in Table \ref{visua_cdd} and Table \ref{visual_levir}. The main targets in the CDD series datasets are scattered single buildings and linear roads, while the main targets in the LEVIR-CD series datasets are dense clusters of buildings. Through comparison, we find that the SSLChange framework improves the capability of geometric structure feature acquisition for CD baselines under both data distributions. In the case of limited data, SSLChange is able to obtain more satisfied feature segmentation results than the original baselines. We believe that the proposed SSLChange framework brings benefits to the baseline for two reasons. The application of the Spatial Projector and Predictor in the SSLChange framework helps to maintain and capture accurate spatial features during the pretext tasks. In addition, the Channel branch in SSLChange reduces the dimension of the samples by MLP module, which makes the framework more sensitive to inter-class differences, leading to more accurate segmentation results in the latent space.

For better illustration, we additionally visualize several hard samples with complex scenarios. As illustrated in Fig. \ref{fig:complex}, although the presented samples with significant background change and shadow, our proposed SSLChange is still able to accurately capture the changed regions. 

\subsubsection{Ablation Experiments}
\label{sec:ablation}
In this part, we perform several ablation experiments to evaluate the main components of the proposed SSLChange framework. The ablation study contains five main experiments: 

\noindent \textbf{Domain Adapter:}
To evaluate the performance of the Domain Adapter (DA), the SSLChange framework is first trained without DA and with DA, respectively. Then the pre-trained encoder is clipped and aligned with the baseline to perform downstream fine-tuning. Specifically, in the case where the framework does not apply DA, the input samples $x$ follow the random data augmentation method in SimSiam \cite{chen2021exploring}, and are sent into the Hierarchical Contrastive Head. We select SNUNet-CD as the baseline for comparison on the LEVIR-CD-50\% dataset. The results of the ablation study are shown in Table \ref{tab:ablation of DA}. \textit{w/o Domain Adapter} represents without Doamin Adapter, and \textit{w/ Domain Adapter} represents with Doamin Adapter.The performance of F1 and IoU metrics demonstrates the superiority of the proposed DA. The adaptive translation between bi-temporal image domains promotes the encoder to capture cross-domain features and eliminate the effects of imaging conditions.

\begin{table}[!htbp]
	\centering
	\caption{Ablation experiments of Domain Adapter on LEVIR-CD-50$\%$ dataset.}
	\label{tab:ablation of DA}
	\renewcommand{\arraystretch}{1.5}
	\setlength{\tabcolsep}{13pt}
	\footnotesize
	{
		\begin{tabular}{c|c|c|c}
			
			\Xhline{1pt}
			\textbf{Baseline} & \textbf{Method} & \textbf{F1} & \textbf{IoU}
			\\ \Xhline{0.8pt}
			\multirow{2}{*}{SSLChange} & w/o Domain Adapter & 74.23 & 59.02
			\\ \cline{2-4}
			 & w/ Domain Adapter & 75.18  & 60.23
			\\ \Xhline{1pt}
			
		\end{tabular}
	}
\end{table}

\noindent \textbf{Encoder Clipping:} In the SSLChange framework, the pre-trained encoder is frozen and clipped, only first 3 layers are preserved. Likewise, we explore the effect on pre-trained encoder clipping. From the results shown in Table \ref{tab:ablation of clipping}, we observe a significant improvement in the performance of SSLChange framework with encoder clipping, which corroborates the feature visualization results in Fig. \ref{features}. It demonstrates that shallow pre-trained features from pixel-level SSL pretext tasks are more instructive for down-stream fine-tuning. 

\begin{table}[!htp]
	\centering
	\caption{Ablation experiments of pre-trained encoder clipping on LEVIR-50$\%$ dataset. }
	\label{tab:ablation of clipping}
	\renewcommand{\arraystretch}{1.5}
	\setlength{\tabcolsep}{12pt}
	\footnotesize
	{
		\begin{tabular}{c|c|c|c}
			
			\Xhline{1pt}
			\textbf{Baseline} & \textbf{Method} & \textbf{F1} & \textbf{IoU}
			\\ \Xhline{0.8pt}
			\multirow{2}{*}{SSLChange} & w/o Encoder Clipping & 72.86 & 57.31
			\\ \cline{2-4}
			 & w/ Encoder Clipping & 75.18  & 60.23
			\\ \Xhline{1pt}
			
		\end{tabular}
	}
\end{table}

\noindent \textbf{Hierarchical Contrastive Head:} The ablation study on the Spatial Projector $\&$ Predictor and the Channel Projector $\&$ Predictor in the Hierarchical Contrastive Head is conducted in this section. Similarly, we select SNUNet-CD as the baseline for comparison on the LEVIR-CD-50\% dataset. The results of the ablation study are shown in Table \ref{tab:ablation of proj and pred}, where $Spa.$ represents Spatial Projector\& Predictor, and $Cha.$ is Channel Projector\& Predictor.
	
\vspace{-0.3cm}
\begin{table}[!hp]
	\centering
	\caption{Ablation experiments of Channel and Spatial Projector $\&$ Predictor on LEVIR-CD-50$\%$ dataset. }
	\label{tab:ablation of proj and pred}
	\renewcommand{\arraystretch}{1.5}
	\setlength{\tabcolsep}{12pt}
	\footnotesize
	{
		\begin{tabular}{c|c|c|c|c}
			
			\Xhline{1pt}
			\textbf{Baseline} & \textbf{+ Spa.} & \textbf{+ Cha.} & \textbf{F1} & \textbf{IoU} 
			\\ \cline{1-5}
			 \multirow{4}{*}{SNUNet-CD} & \ding{55} & \ding{55} & 72.90 & 57.35 
			\\ \cline{2-5}
			  & \ding{55} & \ding{51} & 73.49 & 58.09 
			\\ \cline{2-5}
			  & \ding{51} & \ding{55} & 74.10  & 58.86
			\\ \cline{2-5}
			  & \ding{51} & \ding{51} & 75.18  & 60.23
			\\ \Xhline{1pt}
			
		\end{tabular}
	}
\end{table}
It is worth noting that in the case where only Channel branch is applied, the method degenerates to SimSiam \cite{chen2021exploring} as a comparison. The applications of both Spatial and Channel modules in SSLChange pre-training are proven to provide considerable benefits to the baseline. The Spatial branch performs identical convolution operations to maintain the dimension of the features, while the Channel branch obtains semantic information by feature reduction.

\noindent \textbf{Down-stream Fusion:} For the down-stream fine-tuning in the SSLChange framework, the concatenation operation is applied to fuse the original embedding and the pre-trained features. Here, we further exploit the impact of several fusion operations (i.e. Add, Multiply, Deconvolution and Concatenation) for the down-stream fine-tuning, without significantly increasing the whole model parameters. 

\vspace{0.2cm}
The experimental results demonstrate that the Concatenation fusion utilized in SSLChange shows significant superiority over the Multiply (F1: 71.60\%, IoU: 55.76\%) and the Deconvolution (F1: 65.58\%, IoU: 48.78\%) methods, obtaining a performance of 75.18\% and 60.23\% on F1 and IoU metrics, respectively. When compared to the Add method, the Concatenation method possesses a slight advantage (F1: 75.13\%, IoU: 60.17\%). We argue that the Add fusion assigns equal weights to the two components for fusion, and the Concatenation fusion preserves all feature components and allocates the weights automatically through the subsequent linear layer. Therefore, we select the Concatenation fusion in our implementation.



\begin{figure}[!hbp]
	\centering
	\includegraphics[width=0.95\linewidth]{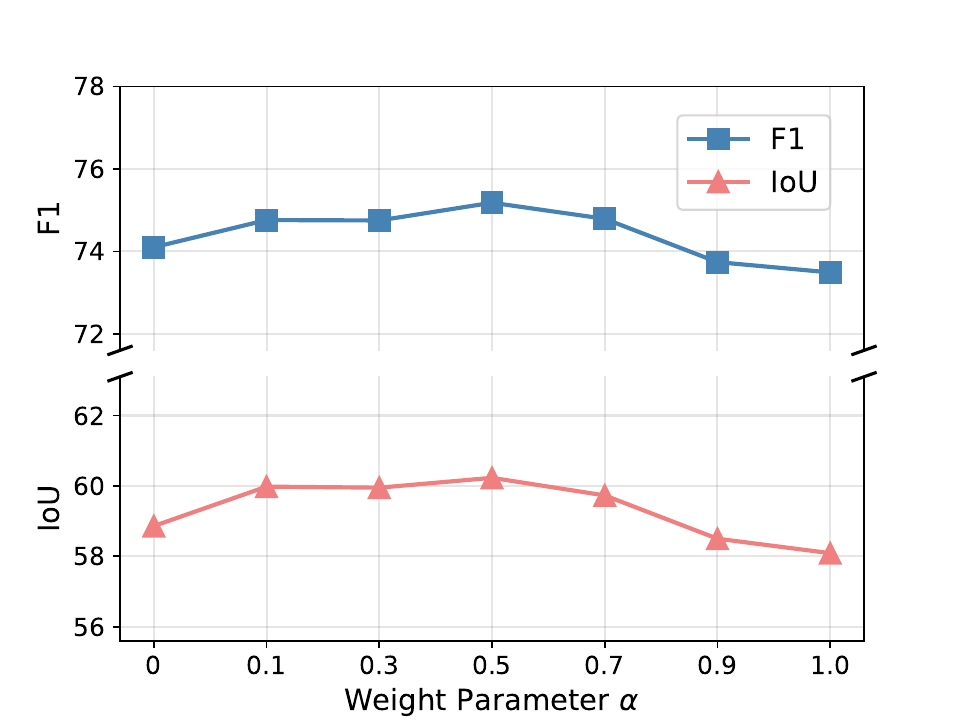}
	\caption{Performance of F1 and IoU under different weight parameter $\alpha$.}
	\label{fig:ablation_alpha}
\end{figure}

\noindent \textbf{Weight Parameter $\alpha$ for Loss Function:} The result curves on the impact of parameter $\alpha$ is presented in Figure \ref{fig:ablation_alpha}. The changing trend reflects that the proposed method achieves the optimal performance when the spatial loss and channel loss reach a relative balance ($\alpha=0.5$). It could be concluded that when the local details from spatial branch are in a equilibrium constraint with the global semantic codes from channel branch, the encoder and the contrastive head are motivated to accurately capture valuable feature representations.

\subsubsection{Cross-Dataset Validation}
To evaluate the generalization ability of the proposed SSLChange framework, we perform cross-dataset validation between the two selected datasets. Considering that only a few labeled data are available in practical scenarios, we transfer the feature representations from a large-scale pre-training dataset to a small-scale downstream dataset. Specifically, we select the CDD dataset as the benchmark for the pre-training of SSLChange. Then the down-stream CD baseline is fine-tuned with the LEVIR-CD series datasets. The experimental results are shown in Table \ref{tab:cross-dataset}, where \textit{Un. LEVIR-CD} and \textit{Un. CDD} represents SSLChange pre-training with unlabeled LEVIR-CD and unlabeled CDD datasets, respectively. In spite of the large differences between the downstream dataset (LEVIR-CD series) and the pre-trained CDD dataset, transferring the CDD pre-trained model to the LEVIR-CD dataset still achieves satisfying performance, which is comparable to the LEVIR-CD pre-training results. Therefore, the generalization capability of the proposed SSLChange framework is further validated through the cross-dataset validation experiments.
\begin{table}[!htbp]
    \centering
    \caption{Cross-Dataset Validation from large-scale pre-training dataset to small-scale downstream dataset.}
    \renewcommand{\arraystretch}{1.7}
    \setlength{\tabcolsep}{2.4pt}
    \footnotesize
    \begin{tabular}{c|cc|cc|cc|cc}

        \Xhline{1pt}
        \multicolumn{1}{c|}{\multirow{2}[0]{*}{\textbf{Pre. Paradigm}}} & \multicolumn{2}{c|}{\textbf{LEVIR-10\%}} & \multicolumn{2}{c|}{\textbf{LEVIR-20\%}} & \multicolumn{2}{c|}{\textbf{LEVIR-50\%}} & \multicolumn{2}{c}{\textbf{LEVIR}} 
        \\
        & \multicolumn{1}{c}{\textbf{F1}} & \multicolumn{1}{c|}{\textbf{IoU}} & \multicolumn{1}{c}{\textbf{F1}} & \multicolumn{1}{c|}{\textbf{IoU}} & \multicolumn{1}{c}{\textbf{F1}} & \multicolumn{1}{c|}{\textbf{IoU}} & \multicolumn{1}{c}{\textbf{F1}} & \multicolumn{1}{c}{\textbf{IoU}} 
        \\ \hline
        Un. LEVIR-CD & 63.92 & 46.97 & 65.68 & 48.89 & 75.18 & 60.23 & 80.25 & 67.01 
        \\ \cline{1-9}
        Un. CDD & 59.71 & 42.56 & 69.26 & 52.97 & 76.31 & 61.70 &  79.58 &  66.09 
        \\ \Xhline{1pt}
    
    \end{tabular}%
  \label{tab:cross-dataset}%
\end{table}%

\subsubsection{Computational Efficiency and Stability}
Computational efficiency and stability are also essential attributes of the proposed SSLChange framework.

\noindent \textbf{Pre-training Time Cost of SSLChange:} 
To get a trade-off between data volume and model performance, we select a mini-batch of 8 to conduct pre-training with the SSLChange framework. We record the pre-training time cost for 100 epochs on two entire datasets: CDD and LEVIR-CD dataset in Table \ref{tab:time-cost}. The proposed SSLChange framework is able to accomplish pre-training with high efficiency regardless of the scale of the dataset. We observe that the SSLChange framework presents a higher pre-training speed on the LEVID-CD dataset than that on the CDD dataset, which is most probably related to limited computing resources. 
\begin{table}[!htp]
    \caption{Pre-training Time Cost of the Proposed SSLChange Framework}
    \label{tab:time-cost}
    \footnotesize
    \centering
    \renewcommand{\arraystretch}{1.7}
    \setlength{\tabcolsep}{4pt}
    \begin{tabular}{c|c|c|c|c}
         \Xhline{1pt}
         \textbf{Dataset} & \textbf{Pre. Batch Size} & \textbf{Data Volume} & \textbf{Total Time} & \textbf{Speed}
         \\ \hline
         CDD & \multirow{2}{*}{8} & 10000 & 6.25 h & 44 FPS
         \\ \cline{1-1} \cline{3-5}
         LEVIR-CD & & 445 & 13.3 min & 56 FPS
         \\ \Xhline{1pt}
    \end{tabular}
\end{table}

\begin{figure}[!htbp]
	\centering
	\includegraphics[width=0.94\linewidth]{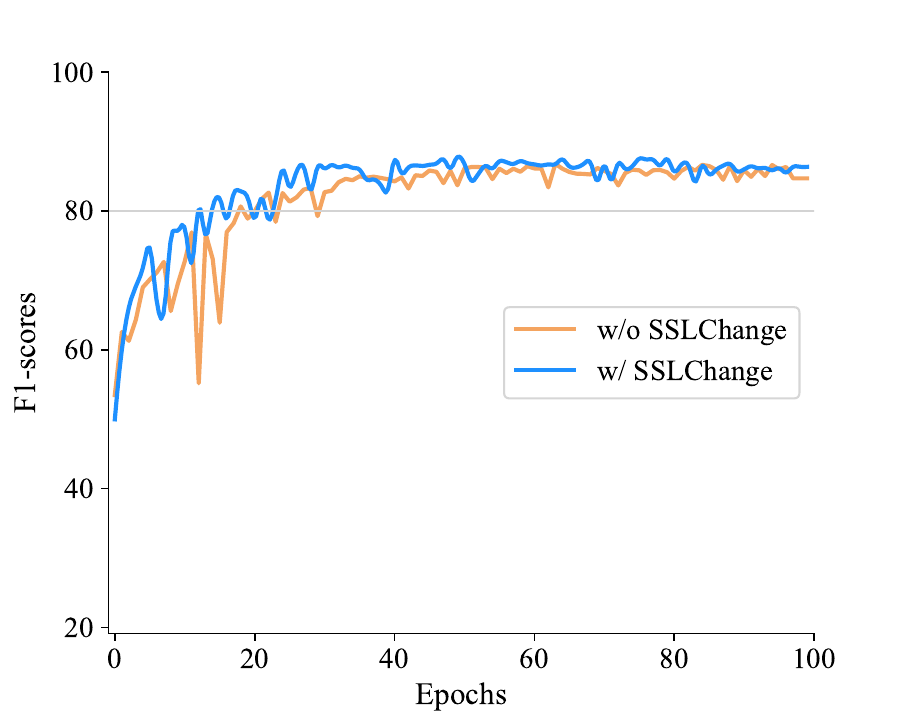}
	\caption{Computational stability and convergence speed of the proposed method during the training process as F1-Epoch curves.}
	\label{F1-Epoch}
\end{figure}

\noindent \textbf{Down-stream Fine-tuning Stability:}
From the curves shown in Fig. \ref{F1-Epoch}, it can be clearly observed that, in the downstream fine-tuning, the training process of the baseline with SSLChange applied is more stable within a limited number of epochs. From the amplitude of the curves, it can also be discovered that SSLChange promotes the baseline to converge faster to the optimal value.

\subsection{Limitation Analysis}
In the SSLChange framework, a domain adapter and a hierarchical contrastive head are utilized to improve the performance of change detection between bi-temporal optical images with self-supervised contrastive pre-training. The experimental results demonstrate the ability to stably enhance the CD baseline in the data-limited situation. However, several existing limitations as follows require further exploration, which also enlighten the future research directions. 

\vspace{-0.5cm}
\subsubsection{Pre-training Paradigm}
Self-supervised contrastive paradigm is adopted in SSLChange which motivates to obtain high-level semantic features. In contrast, self-supervised mask modeling paradigm can promote the model to capture low-level local features by random mask reconstruction. Thus, the integration of the hierarchical feature extraction capabilities from different paradigms in CD task is required to be explored.

\subsubsection{Multi-source Domain Adaption}
The SSLChange framework focuses on the domain adaption between the same source data. However, current airborne and spaceborne platforms are commonly equipped with multi-sensors to jointly acquire data for complementary purpose. Therefore, the generalized domain adaption between multi-source data (e.g., SAR, LiDAR, Text information) remain to be investigated to handle the various data emerging in remote sensing field. 

\subsubsection{Robustness to Pseudo Change}
Some environmental factors and the noises could disturb the model performance, such as seasonal background variations, label noise, unalignment noise, and object occlusion. These factors are also expected to be considered in future research. 

\section{Conclusion}
In this paper, we propose a SSLChange framework for bi-temporal CD tasks, which can be flexibly adapted to existing downstream CD baselines with a concise structure. Specifically, the transferred view is first generated based on the Domain Adapter with single-temporal samples, and then the Spatial branch and the Channel branch in the Hierarchical Contrastive Head are assigned to extract spatial and semantic features, respectively. The pre-trained encoder possesses the ability to accurately capture geometric and categorical information of the targets, which can be directly used for downstream CD baseline fine-tuning after model clipping and alignment. The experimental results demonstrate that our SSLChange framework provides benefits to the baseline in the RS CD task, especially in the case of the data-limited situation enabling the baseline to be more stabilized in the training process and obtaining a more satisfactory performance. 

In future work, we will further consider the existing limitations and explore various self-supervision-based pre-training strategies. In addition, the efficient migration of pre-training generalized frameworks to downstream tasks at the cost of smaller computational resources is another direction of our research.

\small
\bibliographystyle{IEEEtranN}

\end{document}